\documentclass[journal]{IEEEtran}
\usepackage{amsmath,amsfonts}
\usepackage{algorithmic}
\usepackage{array}
\usepackage[caption=false,font=normalsize,labelfont=sf,textfont=sf]{subfig}
\usepackage{textcomp}
\usepackage{stfloats}
\usepackage{url}
\usepackage{verbatim}
\usepackage{graphicx}
\def\BibTeX{{\rm B\kern-.05em{\sc i\kern-.025em b}\kern-.08em
    T\kern-.1667em\lower.7ex\hbox{E}\kern-.125emX}}
\usepackage{balance}

\usepackage{hyperref} 
\usepackage{multirow}
\usepackage{xspace}
\usepackage{booktabs}
\usepackage[capitalise]{cleveref}

\hypersetup{
pdfauthor={Luca Della Libera, Pooneh Mousavi, Salah Zaiem, Cem Subakan, Mirco Ravanelli},
pdftitle={CL-MASR: A Continual Learning Benchmark for Multilingual ASR},
pdfsubject={Continual learning},
pdfkeywords={Continual learning, multilingual ASR, benchmark}
}

\usepackage{xr-hyper} 
\makeatletter

\newcommand*{\addFileDependency}[1]{% argument=file name and extension
\typeout{(#1)}% latexmk will find this if $recorder=0
\@addtofilelist{#1}
\IfFileExists{#1}{}{\typeout{No file #1.}}
}\makeatother

\makeatletter
\DeclareRobustCommand\onedot{\futurelet\@let@token\@onedot}
\def\@onedot{\ifx\@let@token.\else.\null\fi\xspace}
 
\def\ie{{i.e}\onedot} 
 
\def\etc{{etc}\onedot}

\def\etal{{et al}\onedot}
\makeatother

\begin{document}

\title{CL-MASR: A Continual Learning Benchmark for Multilingual ASR}

\author{Luca Della Libera$^{*}$, Pooneh Mousavi$^{*}$, Salah Zaiem, Cem Subakan,  Mirco Ravanelli
\thanks{$^{*}$Equal contribution.}
\thanks{L. Della Libera, P. Mousavi, and  M. Ravanelli are with the Gina Cody School of Engineering and Computer Science, Concordia University, Montreal, Canada. (e-mail: luca.dellalibera@mail.concordia.ca; pooneh.mousavi@mail.concordia.ca; mirco.ravanelli@mail.concordia.ca).}
\thanks{S. Zaiem is with LTCI, Télécom Paris, Paris, France. (e-mail: salah.zaiem@telecom-paris.fr).}
\thanks{C. Subakan is with the Department of Computer Science and Software Engineering, Université Laval, Québec, Canada. (e-mail: cem.subakan@ift.ulaval.ca).}}

%\markboth{IEEE/ACM Transactions on Audio, Speech, and Language Processing,~Vol.~18, No.~31, October~2023}%
%{CL-MASR: A Continual Learning Benchmark for Multilingual ASR}

\maketitle

\begin{abstract}
Modern multilingual automatic speech recognition (ASR) systems like Whisper have made it possible to transcribe audio in multiple languages with a single model. However, current state-of-the-art ASR models are typically evaluated on individual languages or in a multi-task setting, overlooking the challenge of continually learning new languages. There is insufficient research on how to add new languages without losing valuable information from previous data. Furthermore, existing continual learning benchmarks focus mostly on vision and language tasks, leaving continual learning for multilingual ASR largely unexplored.
To bridge this gap, we propose CL-MASR, a benchmark designed for studying multilingual ASR in a continual learning setting. CL-MASR provides a diverse set of continual learning methods implemented on top of large-scale pretrained ASR models, along with common metrics to assess the effectiveness of learning new languages while addressing the issue of catastrophic forgetting. 
To the best of our knowledge, CL-MASR is the first continual learning benchmark for the multilingual ASR task. The code is available at \href{https://github.com/speechbrain/benchmarks}{https://github.com/speechbrain/benchmarks}.
\end{abstract}


\section{Introduction}
\label{sec:introduction}
\IEEEPARstart{A}{utomatic} speech recognition (ASR) has traditionally focused on converting speech into written text for individual languages. However, with the increasing need for cross-lingual communication and the availability of large-scale multilingual datasets, the focus has recently shifted towards the development of massively multilingual ASR models. These models leverage the well-known advantages of scaling~\cite{kaplan2020scaling} and effectively use similarity across languages to improve performance.
Notably, the emergence of models like M-CTC-T~\cite{Lugosch2021PseudoLabelingFM}, Whisper~\cite{radford2022robust}, USM~\cite{zhang2023google}, and MMS~\cite{pratap2023mms} has enabled the automatic transcription of audio of hundreds of languages using a single shared model. Despite these remarkable achievements, with over 7,000 languages existing worldwide~\cite{lewis2016ethnologue}, the problem of multilingual ASR is far from fully solved. Indeed, due the dynamic nature of language,
even the most advanced ASR system would need regular updates to effectively deal with new dialects and/or domain-specific lexicons, or other variations of existing languages.

The problem of adapting a model over time is addressed by the field of continual learning (CL), also referred to as lifelong learning or incremental learning. It focuses on designing algorithms that continuously learn from a sequence of tasks.

A naive CL approach is to fine-tune the ASR model on data from new languages as they become available. Unfortunately, this often results in \textit{catastrophic forgetting}~\cite{forgetting_french, goodfellow2013empirical}, a phenomenon that occurs when adjusting the model's weights based on data from a different distribution{. This distribution shift} hampers the model's performance on previously learned tasks.
Various methods have been proposed to mitigate catastrophic forgetting in supervised learning, including rehearsal-based~\cite{rolnick2019experience,chaudhry2019efficient,replay_shin,icarl_rebuffi}, architecture-based~\cite{pnn_rusu,mallya2018piggyback,expert_aljundi,packnet_mallya}, and regularization-based~\cite{ewc_kirkpatrick,zenke2017continual,hou2019learning} {approaches}. While these techniques have been extensively explored in vision and text domains for knowledge transfer across tasks, ASR has received limited attention, especially in multilingual settings.

One of the few works in this specific area is the one by Li \etal~\cite{cl-asr_li}, who conduct an experimental study to analyze the impact of model's capacity when incorporating additional languages in a massively multilingual ASR model.
However, their investigation solely focuses on the application of incremental fine-tuning, thereby leaving room for exploring a wide range of CL techniques. Additionally, all their models are trained \emph{from scratch}.
Research by Ostapenko \etal~\cite{ostapenko2022foundational} emphasizes the importance of employing large-scale pretrained models in the fields of computer vision and natural language processing (NLP) for CL. When training a model from scratch for CL, high-level features are often unstable. On the contrary, using large-scale pretrained models can yield to more robust and versatile hidden representations that could better generalize to newly introduced tasks. There are two types of pretrained models that are commonly used in ASR. The first consists of large-scale \emph{supervised} models like Whisper~\cite{radford2022robust}, which encompass both an encoder and a decoder.
The second includes \emph{self-supervised} models such as wav2vec 2.0~\cite{baevski2020wav2vec}, HuBERT~\cite{hsu2021hubert}, and WavLM~\cite{chen2022wavlm}, which are employed as encoder-only models for audio feature extraction.
Given the potential differences in training dynamics when applied to CL, it is important to explore the use of both types of models.

To address the lack of research {in} exploring the potential of large-scale pretrained multilingual ASR models for CL, as well as the scarcity of diverse CL methods specifically designed for multilingual ASR, we introduce \textbf{C}ontinual \textbf{L}earning for \textbf{M}ultilingual \textbf{ASR} (CL-MASR), a benchmark for continual learning applied to multilingual ASR.
CL-MASR provides:
\begin{itemize}
    \setlength\itemsep{.001cm}
    \item A curated selection of challenging medium/low-resource languages from the Common Voice 13~\cite{ardila2020common} dataset to utilize for CL experiments in the context of multilingual ASR.
    \item A diverse set of well-known CL methods implemented on top of Whisper~\cite{radford2022robust} and WavLM~\cite{chen2022wavlm}, along with standard evaluation metrics.
    \item A modular platform based on the popular SpeechBrain~\cite{speechbrain_ravanelli} toolkit that enables the easy extension to other CL strategies and/or pretrained models, and facilitates analysis and visualization of the experimental results. Our goal is to provide a friendly interface for CL researchers in order for them to easily test their novel methods on CL-MASR.
\end{itemize}
Based on the experiments conducted using CL-MASR, we conclude that experience replay~\cite{rolnick2019experience} is on average one of the best performing CL approaches in the context of multilingual ASR.

\begin{figure*}[t]
  \centering
  \includegraphics[width=1.0\textwidth]{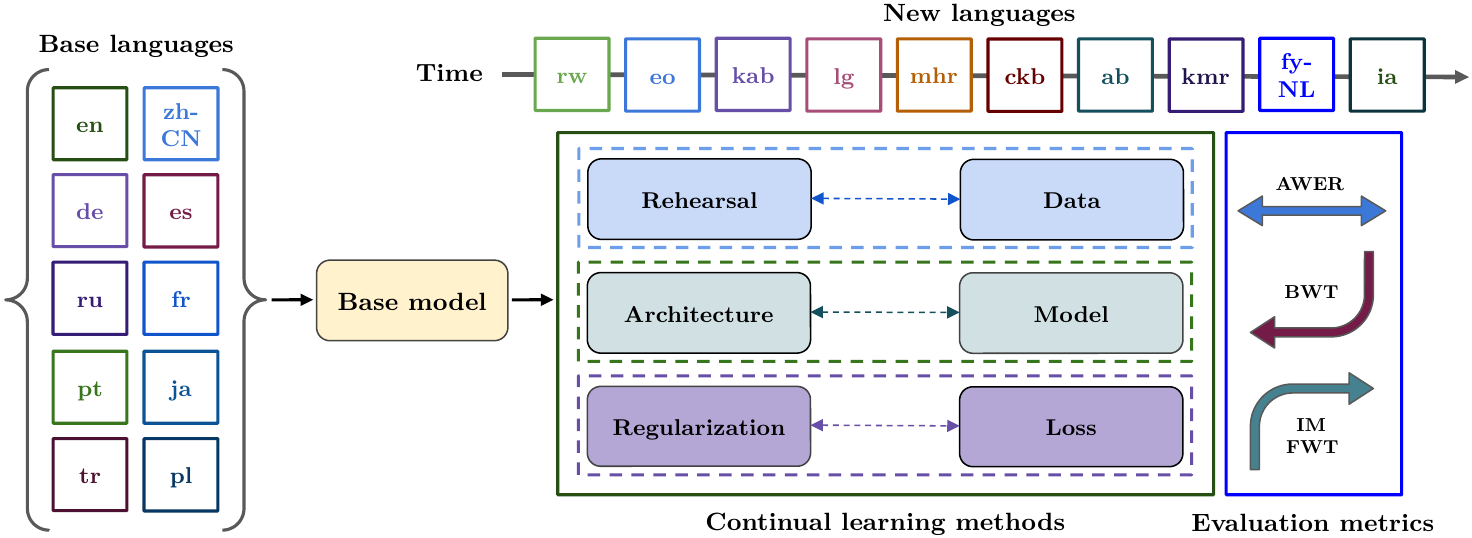}
  \caption{The workflow of CL-MASR consists of initial pretraining of a base model using a set of base languages. New languages are added incrementally in the same order as depicted in the figure. Rehearsal-based methods operate on the data, architecture-based on the model, and regularization-based on the loss function.
  Average word error rate (AWER), backward transfer (BWT), intransigence measure (IM), and forward transfer (FWT) are used to assess the performance.}
  \label{fig:pipleline}
\end{figure*}

\section{Related Work}
\textbf{Multilingual ASR.} In recent years, we have witnessed impressive progress in the field of multilingual ASR, with the development of massively multilingual models capable of transcribing tens or even hundreds of languages. One such model is M-CTC-T~\cite{Lugosch2021PseudoLabelingFM}, a transformer encoder architecture with a connectionist temporal classification (CTC)~\cite{graves2006connectionist} head trained using a combination of supervised and semi-supervised techniques on Common Voice 6~\cite{ardila2020common} and VoxPopuli~\cite{wang-etal-2021-voxpopuli} datasets. Notably, it takes advantage of unlabeled data to improve its performance and supports 60 different languages. Another example is Whisper~\cite{radford2022robust}, a transformer-based encoder-decoder model trained in a \emph{weakly} supervised manner using noisy data crawled from the web. It is designed to perform ASR as well as voice activity detection, language identification, speech translation, and timestamp prediction. Trained on 680,000 hours of annotated speech, it can successfully transcribe 99 different languages without requiring any fine-tuning.

\textbf{Continual Learning in Vision and Text Domains.}
CL has been extensively explored in both vision and text~\cite{wang2023comprehensive,de2021continual,ke2022continual,mai2022online, antoniou2020defining, ke2022continual, lomonaco2021avalanche, masana2022class,mai2022online}. Several datasets and benchmarks exist for studying CL in the visual domain. For example, permuted MNIST~\cite{goodfellow2013empirical}, CORe50~\cite{lomonaco2017core50}, Split-MNIST~\cite{zenke2017continual}, and Split-CIFAR~\cite{krizhevsky2009learning,icarl_rebuffi} employ synthetic data to evaluate various CL methods for image classification tasks. Visual Domain Decathlon~\cite{rebuffi2017learning} and  CLEAR~\cite{lin2021clear}, on the contrary, propose CL benchmarks based on real-world images. In a more recent study by Bornschein \etal~\cite{bornschein2022nevis}, a novel benchmark comprising more than 100 visual classification tasks is introduced. CL is also an active research area in NLP. CLIF~\cite{jin2021learn} and Natural Language Decathlon~\cite{mccann2018natural} are benchmarks for CL in NLP tasks including entity typing, sentiment analysis, and natural language inference.
Other works~\cite{greco2021measuring,srinivasan2022climb} explore CL methods for multimodal tasks at the intersection of vision and language such as visual question answering. 
These studies demonstrate the benefits of CL approaches for learning new tasks while avoiding catastrophic forgetting in diverse domains. Their findings motivate us to investigate the effectiveness of CL applied to multilingual ASR.

\textbf{Continual Learning in ASR.}
Most of the research on CL in the ASR domain revolves around domain-incremental learning (DIL)~\cite{van2022three}, where tasks share the same data label space but come from different distributions. Sadhu \etal~\cite{sadhu2020continual} propose a method in which the likelihood of a monolingual HMM-DNN ASR model is decomposed into sub-models  specific for each data domain. Chang \etal~\cite{chang2021towards} extend a monolingual hybrid CTC-transformer model to new data distributions. Li \etal~\cite{li2023efficient} leverage the self-supervised pretrained BEST-RQ~\cite{chiu2022self} and JUST hydra~\cite{hwang2022pseudo} joint training strategy for fine-tuning using both source and target domain data. These studies focus primarily on examining the effects of CL methods in DIL scenarios, leaving room for further exploration of CL techniques in settings such as multilingual ASR, where the data label space for each task may differ~\cite{van2022three}. Relatedly, limited attention has been devoted to CL for multilingual ASR. Li \etal~\cite{cl-asr_li} analyze how the capacity of a massively multilingual ASR model influences its performance when integrating additional languages. Kessler \etal~\cite{kessler2021continual} employ a pretrained wav2vec 2.0~\cite{baevski2020wav2vec} model to address the challenge of continually learning new language representations. However, there is a lack of research on the benefits of applying CL methods to expand the capabilities of large pretrained ASR models like Whisper~\cite{radford2022robust}. In this work, our goal is to introduce the first CL benchmark for multilingual ASR that incorporates large pretrained models.
\section{Problem Formulation}
CL focuses on the sequential learning of a model for multiple tasks while preserving the knowledge obtained from previous tasks. The primary assumption is that we have no or limited access to data from previous tasks. The goal is to learn new tasks without experiencing catastrophic forgetting.
CL techniques can be categorized into different groups based on their intended use. One such category is task-incremental learning (TIL)~\cite{wang2023comprehensive}, which is well-suited for scenarios where tasks have distinct data label spaces and the identity of the task is provided during both training and inference phases. This is the case, for example, of indoor vs outdoor scene classification.
On the other hand, blurred boundary continual learning (BBCL)~\cite{wang2023comprehensive} is specifically designed for situations where task boundaries are unclear, and data label spaces are not entirely disjoint. For instance, in reinforcement learning, it is common for the agent to experience a gradual change of tasks as it takes actions.
Multilingual ASR lies at the intersection of TIL and BBCL, as some labels (\ie tokens) are shared across multiple languages, while others are specific to certain languages.
This similarity among languages requires the utilization of approaches that can handle both distinct and overlapping data labels effectively.

In this work, we consider the learning of each language as a distinct task.
Formally, let $\theta_0$ denote the parameters of a \textbf{base model} that can transcribe speech from a set of $L_\text{base}$ 
%\cem{number of}
\textbf{base languages} and $\mathcal{D_\text{new}}= \cup_{i=1}^{L_\text{new}}\{(X_i, Y_i)\}$ a dataset of $L_\text{new}$ 
%\cem{number of} 
\textbf{new languages} with $X_i$ representing speech samples from the $i$-th new language and $Y_i$ the corresponding transcriptions.
We aim to incrementally train a sequence of models $\theta_1, \dots, \theta_{L_\text{new}}$ on $(X_1, Y_1), \dots, (X_{L_\text{new}}, Y_{L_\text{new}})$ where the $i$-th model can successfully transcribe all the languages up to $i$-th, including the base ones.
The main challenge is to prevent each model from forgetting the previously learned languages.

\section{Benchmark Design}
\label{sec:design}
We introduce CL-MASR, the first benchmark for CL in multilingual ASR. It includes the following four components: a dataset of speech-transcription pairs from multiple languages, a selection of base models to train incrementally, a variety of CL methods along with their implementations, and a standard set of evaluation metrics. The workflow is depicted in \cref{fig:pipleline}.
The benchmark platform, based on the popular SpeechBrain~\cite{speechbrain_ravanelli} toolkit, is available at \href{https://github.com/speechbrain/benchmarks}{https://github.com/speechbrain/benchmarks} and is licensed under Apache 2.0\footnote{\href{https://www.apache.org/licenses/LICENSE-2.0}{https://www.apache.org/licenses/LICENSE-2.0}}.

\subsection{Dataset}
\label{subsec:dataset}
Our benchmark builds upon Common Voice 13\footnote{\href{https://commonvoice.mozilla.org/en}{https://commonvoice.mozilla.org/en}}~\cite{ardila2020common}. This public dataset, obtained through crowd-sourcing, consists of short audio recordings and their corresponding transcriptions for 108 languages, comprising a total of 17,690 validated hours divided into training, validation, and test splits. We select from it the following two sets of languages ($L_\text{base} = L_\text{new} = 10$):
\begin{itemize}
    \item \textbf{base languages}: English (en), Chinese (zh-CN), German (de), Spanish (es), Russian (ru), French (fr), Portuguese (pt), Japanese (ja), Turkish (tr), and Polish (pl).
    \item \textbf{new languages}: Kinyarwanda (rw), Esperanto (eo), Kabyle (kab), Luganda (lg), Meadow Mari (mhr), Central Kurdish (ckb), Abkhaz (ab), Kurmanji Kurdish (kmr), Frisian (fy-NL), and Interlingua (ia).
\end{itemize}
The languages in the first group have a substantial amount of data available on the web and are typically well-supported by multilingual ASR systems~\cite{Lugosch2021PseudoLabelingFM,radford2022robust,zhang2023google,pratap2023mms}. On the other hand, the second group consists of medium/low-resource languages that are often overlooked in ASR research. These languages have limited data available, making them more challenging to work with compared to the base languages. For example, Whisper~\cite{radford2019language} cannot transcribe any of them.
For each language, we randomly extract up to 10 hours of data for training, 1 for validation, and 1 for testing, respectively.
This approach not only reduces the computational burden when fine-tuning large multilingual ASR models, but also reflects a more realistic scenario where very limited data are available when incrementally learning new tasks.
In order to improve the quality and consistency of the data, we apply some minimal preprocessing steps. First, we filter out utterances longer than 10 seconds, as they are likely to be recordings from open microphones or contain excessive background noise. Then, we discard utterances with transcriptions longer than 200 characters to limit memory usage. Additionally, we perform basic transcript normalization by removing punctuation marks and collapsing repeated spaces into a single one. For detailed information about the data distribution, refer to \cref{tab:cv}.

\begin{table}[t]
  \caption{Languages from Common Voice 13~~\cite{ardila2020common} used in our benchmark.
  Reported values refer to the raw subsampled data before applying any preprocessing.}
  \label{tab:cv}
  \centering
    \begin{tabular}{lc@{\qquad}ccc}
  \toprule
  \multirow{2}{*}{\raisebox{-\heavyrulewidth}{Language}} & \multirow{2}{*}{\raisebox{-\heavyrulewidth}{ISO 639-1}} & \multicolumn{3}{c}{Duration (minutes)}\\
  \cmidrule{3-5}
  & & Training & Validation & Test \\
  \midrule
 \multicolumn{5}{c}{Base languages}\\
 \midrule
  English & en & 600 & 60 & 60 \\
  Chinese & zh-CN & 600 & 60 & 60 \\
  German & de & 600 & 60 & 60 \\
  Spanish & es & 600 & 60 & 60 \\ Russian & ru & 600 & 60 & 60 \\ French & fr & 600 & 60 & 60 \\ Portuguese  & pt & 600 & 60 & 60 \\ Japanese & ja & 586 & 60 & 60 \\ Turkish & tr & 600 & 60 & 60 \\ Polish & pl & 600 & 60 & 60 \\
  \midrule
 \multicolumn{5}{c}{New languages}\\
 \midrule
  Kinyarwanda & rw & 600 & 60 & 60 \\
  Esperanto & eo & 600 & 60 & 60 \\
  Kabyle & kab & 600 & 60 & 60 \\
  Luganda & lg & 600 & 60 & 60 \\
  Meadow Mari & mhr & 600 & 60 & 60 \\
  Central Kurdish & ckb & 484 & 60 & 60 \\
  Abkhaz & ab & 600 & 60 & 60 \\
   Kurmanji Kurdish & kmr & 296 & 60 & 60 \\
  Frisian & fy-NL & 330 & 60 & 60 \\
  Interlingua & ia & 313 & 60 & 60 \\  
  \bottomrule
\end{tabular}
\end{table}

\subsection{Models}
There are two main types of pretrained models for multilingual ASR: \emph{supervised} and \emph{self-supervised}.
We explore the application of both categories to CL.

As a supervised pretrained ASR model, we employ the large-v2 version of Whisper\footnote{\href{https://huggingface.co/openai/whisper-large-v2}{https://huggingface.co/openai/whisper-large-v2}}~\cite{radford2022robust}.
Based on the encoder-decoder transformer~\cite{vaswani_transformer} architecture, the encoder extracts audio features from input Mel spectrograms. Conditioned on the encoder's hidden representations, the autoregressive decoder generates the corresponding transcriptions in a multi-task format that involves the use of special tokens as task specifiers.
Since Whisper already supports all the base languages, there is no need for extra fine-tuning and can be directly used in a CL fashion to sequentially add new languages.
Note however that CL with Whisper presents additional challenges compared to standard sequence-to-sequence models.
First, the multi-task training format requires predicting a language-specific token at the beginning of each transcript to allow for language identification. Therefore, a new token embedding must be trained for each new language.
Second, since Whisper employs a universal byte-level BPE~\cite{radford2019language} tokenizer, its token space of size 51,865 is shared among all the languages.
We observed that, because of this, Whisper is more susceptible to catastrophic forgetting.
Finally, given the model size of $\sim$1,550 M parameters,
memory is a limiting factor when applying CL strategies.

As a self-supervised pretrained ASR model, we employ
the large version of WavLM\footnote{\href{https://huggingface.co/microsoft/wavlm-large}{https://huggingface.co/microsoft/wavlm-large}}~\cite{chen2022wavlm} to extract audio features followed by two bidirectional LSTM (BiLSTM)~\cite{hochreiter_lstm} layers with a 1,024-dimensional hidden state and a linear projection to the token space. This architecture is widely recognized and has demonstrated reliable performance in the SUPERB~\cite{yang2021superb} benchmark. First of all, we fit a character-level SentencePiece~\cite{kudo-richardson-2018-sentencepiece} tokenizer on the available transcriptions from both the base and the new languages, resulting in a vocabulary of 4,887 tokens. Then, we jointly fine-tune the model on all the base languages via CTC~\cite{graves2006connectionist} loss. We train the model for 20 epochs using the AdamW~\cite{adamw} optimizer with an initial learning rate of 0.0001 and a batch size of 8. After each epoch, the learning rate is reduced by 20\% if no validation performance improvement is observed. We clip the gradient ${L}_2$ norm to 5 to enhance stability. We also freeze the convolutional layers in WavLM's encoder and enable automatic mixed-precision to reduce memory consumption and speed up training.
Greedy decoding is used for inference.
The fine-tuned model obtained from this process has a total of $\sim$367 M parameters and serves as the starting point for the CL experiments with self-supervised multilingual ASR.

\subsection{Continual Learning Methods}
\label{subsect:cl_methods}
CL algorithms can be categorized into three main groups: \emph{rehearsal-based}, \emph{architecture-based}, and \emph{regularization-based} approaches. The first involve storing data from previous tasks or utilizing generative models. The second consist in progressively expanding the model with task-specific sub-networks. The third employ regularization techniques to induce knowledge sharing between tasks and prevent forgetting.
In addition to naive fine-tuning, which provides a lower bound for the overall performance, we experiment with methods from all these categories. In particular, we implement:
\begin{itemize}
    \item \textbf{Fine-tuning (FT)}: we sequentially fine-tune Whisper and WavLM on the new languages via cross-entropy and CTC~\cite{graves2006connectionist} loss, respectively. We train the models
    %we sequentially fine-tune the base model on the new languages
    for 2 epochs per language using the AdamW~\cite{adamw} optimizer with an initial learning rate of 0.0001 and the maximum batch size allowed by our hardware. After each epoch, the learning rate is reduced by 20\% if no validation performance improvement is observed. We clip the gradient ${L}_2$ norm to 5 to enhance stability. We also freeze Whisper's encoder and the convolutional layers in WavLM's encoder and enable automatic mixed-precision to reduce memory consumption and speed up training. Greedy decoding is used for inference. When training Whisper, for each language a new randomly initialized embedding corresponding to the special token for language identification is appended to the decoder's embedding layer and the entire layer is fine-tuned. Note that the embedding matrix is shared with the final linear projection. When testing Whisper on a given language, following~\cite{radford2022robust}, we manually force the correct special token for language identification instead of using the one predicted by the model. Unless otherwise specified, the same training setup is used for all the following methods.
    \item \textbf{Experience replay (ER)}~\cite{rolnick2019experience}: this rehearsal-based method employs a buffer that retains a portion of data from previous tasks. These stored experiences are then replayed when learning about the current task in order to prevent catastrophic forgetting and promote knowledge transfer. ER can be implemented at either the batch level or the dataset level. In our experiments, we found the latter approach to be more effective. Before training, we mix data from the current task with randomly sampled data from each previous task with a replay ratio of 10\%.
    \item \textbf{Averaged gradient episodic memory (A-GEM)}~\cite{chaudhry2019efficient}: similarly to ER, this rehearsal-based method utilizes a replay buffer. However, it differs from ER by treating the losses on previous experiences as inequality constraints, avoiding their increase while allowing their decrease. This approach tries to actively prevent catastrophic forgetting while potentially improving the performance on previous tasks. Specifically, A-GEM calculates the average gradient of the replayed samples. If the mean gradient and the current gradient point in the same direction, the current gradient is used to update the model's parameters. Otherwise, the orthogonal projection to the averaged gradient is used. As in ER, we retain 10\% of the data from each previous task.
    \item \textbf{Dark experience replay (DER)}~\cite{der}: this rehearsal-based method is an extension of vanilla ER that combines the replay of past experiences with knowledge distillation. In particular, it encourages the current task's model (\ie the \emph{student}) to mimic the responses of a previous task's model (\ie the \emph{teacher}) on data from the previous task.
    This is achieved by minimizing the loss on the current task plus the squared ${L}_2$ distance between the student's output logits and the teacher's output logits, computed on the corresponding previous task's experiences sampled from a replay buffer.
    Note that, although we apply this technique in a task-incremental fashion, when paired with reservoir sampling~\cite{reservoir}, it can effectively deal with the challenging scenario where tasks boundaries are blurred and new data flow in continuously.
    The trade-off between mimicking the teacher and accommodating new knowledge from the current task is controlled by the hyperparameter $\alpha$, which determines the relative strength of the regularization term.
    In our experiments, we set \mbox{$\alpha$ = 1} and, as in ER and A-GEM, we retain 10\% of the data from each previous task.
    \item \textbf{Progressive neural networks (PNN)}~\cite{pnn_rusu}:
    this architecture-based method introduces identical sub-networks for each task and allows knowledge transfer among them via lateral adaptor connections. Such a progressive expansion allows the model to learn new tasks while retaining knowledge from previous ones without suffering from catastrophic forgetting. However, PNN requires the task identity to be provided at inference time. In our experiments, we extend Whisper by adding a final task-specific transformer~\cite{vaswani_transformer} decoder layer and a corresponding embedding layer for each language.
    Similarly, for WavLM, we use two BiLSTM~\cite{hochreiter_lstm} layers with a 1,024-dimensional hidden state and a corresponding linear projection. All the other parameters stay frozen during training.
    \item \textbf{Piggyback (PB)}~\cite{mallya2018piggyback}:
    this architecture-based method involves selectively masking the frozen parameters of a base model. This is achieved by maintaining a set of learnable real-valued weights that undergo a deterministic thresholding function, resulting in binary masks. These masks are then applied to the existing parameters of the model. By updating the real-valued weights via gradient descent, the goal is to learn task-specific masks that are well-suited for the given task.
    Not only is this approach immune to catastrophic forgetting, but it is also agnostic to task ordering and memory-efficient as the masks incur a low overhead of only 1 bit per parameter. However, PB requires the task identity to be provided at inference time.
    In our experiments, due to memory constraints, we mask only the last two layers of each model. Furthermore, task-specific embedding layer and linear projection are trained for Whisper and WavLM, respectively.
    The real-valued weights are initialized to 0.01 and the masking threshold is set to 0.005.  
    \item \textbf{Learning to prompt (L2P)}~\cite{l2p_wang}: this method falls under the umbrella of prompt tuning~\cite{prompt-tuning} techniques, which have recently emerged with the proliferation of large language models. However, it can be broadly classified as an architecture-based approach, since a prompt can be seen as a simple task-specific adapter network.
    L2P maintains a pool of \emph{prompts}, \ie learnable vectors, that are selectively injected into a frozen pretrained model to modify its behavior and adapt it to a new task.
    Prompts can be chosen either on a per-instance or per-task basis. Furthermore, they can be incorporated at different stages of the architecture such as the encoder's or decoder's input layer, the encoder's last hidden layer, \etc via concatenation, Hadamard product, or other fusion operations.
    In our experiments, we use the simpler per-task variant. In particular, we maintain a pool of 10 prompts, one for each new language, of shape $d_{\text{model}} \times d_{\text{model}}$, where $d_{\text{model}}$ is the dimensionality of the model's hidden representations, and we post-multiply the encoder's output by the prompt of the corresponding language. Similarly to PNN and PB, the task identity needs to be provided at inference time.
    \item \textbf{Elastic weight consolidation (EWC)}~\cite{ewc_kirkpatrick}: this regularization-based method estimates the importance of each parameter based on its contribution to the performance on previous tasks. It does so by computing the Fisher information matrix (FIM), which measures the sensitivity of the loss function with respect to the model's parameters. Based on this sensitivity, EWC calculates a quadratic penalty term that is added to the loss function while training on the current task. This term encourages the model to preserve the important parameters for previous tasks while adapting to the new task.
    Such a trade-off is explicitly controlled by the hyperparameter $\lambda$, which determines the relative strength of the regularization term.
    Due to memory constraints, we resort to an online version of EWC and include the penalty term only for the last task. To mitigate the effect of this approximation, we introduce an hyperparameter, $\alpha$, to control the influence of previous tasks in updating the parameter sensitivity. In our experiments, we set $\lambda$ = 5 and $\alpha$ = 0.5.
    Also note that the original training data for Whisper are not publicly available. Hence, we estimate the initial FIM using the training data from the base languages.
    \item \textbf{Learning without forgetting (LwF)}~\cite{hou2019learning}:
    this regularization-based method involves using a frozen copy of the model trained up to the previous task as a \emph{teacher} and the current model as a \emph{student}. The goal is to enable the student to learn not only from the current task but also from the teacher.
    To achieve this, LwF utilizes knowledge distillation~\cite{hinton2015distilling}. First, the output probabilities of the teacher are computed for each sample in the current task. Then, the student is trained to minimize the loss on the current task plus the cross-entropy between teacher's and student's output probabilities, smoothed by a temperature parameter $T$ that increases the weight for smaller probabilities.
    The trade-off between mimicking the teacher and accommodating new knowledge from the current task is controlled by the hyperparameter $\lambda$,  which determines the relative strength of the regularization term.
    In our experiments, we set $T$ = 2 and $\lambda$ = 10.
    \item \textbf{Memory Aware Synapses (MAS)}~\cite{mas}: similarly to EWC, this regularization-based method mitigates catastrophic forgetting by penalizing large updates to parameters that contribute the most to the performance on previous tasks. Differently from EWC, it estimates parameter relevance as the average magnitude of the gradients of the squared ${L}_2$ norm of the learned function. Not only is this definition of importance agnostic to the loss function, but it can also be accumulated over tasks in an online manner.
    The trade-off between forgetting and learning is explicitly controlled by the hyperparameters $\lambda$ and $\alpha$, which determine the relative strength of the regularization term and the influence of previous tasks in updating the parameter importance, respectively. In our experiments, we set $\lambda$ = 1 and $\alpha$ = 0.5. Furthermore, as for EWC, we estimate the initial parameter importance using the training data from the base languages.
\end{itemize}

\subsection{Evaluation Metrics}
When evaluating CL methods, a primary consideration is to assess the \emph{overall performance} across all learned tasks. To do so, we employ a variation of the average accuracy~\cite{chaudhry2018riemannian,lopez2017gradient, hou2019learning}, the \textbf{average word error rate (AWER)}\footnote{For Chinese and Japanese the concept of word is not well-defined, hence, following~\cite{radford2022robust}, we consider the character error rate instead of the word error rate.}, calculated incrementally after each newly introduced task:
\begin{equation}
\text{AWER}_{t} = \frac{1}{t}
\sum_{i=1}^t \text{WER}_{t,i} \, \, \, t = 1, \dots, T, \label{eq:awer}
\end{equation}
where $T$ denotes the total number of tasks and $\text{WER}_{t,i}$ {denotes} the word error rate on the $i$-th task after the model finishes the learning on the $t$-th task. In particular, in our benchmark, we consider the $10$ base languages as a single joint task. Hence, since we have $10$ new languages, {we set} $T = 1 + 10 = 11$.
Here, $\text{WER}_{t,1}$ represents the word error rate averaged over the $10$ base languages after learning about the $t$-th task.
{Note that} $\text{AWER}_{t}$ $\in [0, \infty)$, with smaller values indicating better overall performance. It provides a comprehensive understanding of the model's ability to retain and utilize knowledge from previously learned tasks while accommodating new information.

Another key aspect is memory \emph{stability}{, or robustness against \emph{forgetting}}, \ie the impact of learning new tasks on the performance of previously learned tasks. More specifically, we aim to understand whether learning new tasks has any detrimental effect on the model's performance on previously learned ones. To this end, we measure the \textbf{backward transfer (BWT)}~\cite{lopez2017gradient}, defined as
\begin{equation}
\text{BWT}_{t} = \frac{1}{t-1}
\sum_{i=1}^{t-1} \text{WER}_{i,i} - \text{WER}_{t,i} \, \, \, t = 2, \dots, T, \label{eq:bwt}
\end{equation}
\ie the average performance gain on previously learned tasks.
{Note that} $\text{BWT}_{t}$ $\in (-\infty, \infty)$ with positive values indicating improvement on the previously learned tasks and negative ones indicating forgetting. However, $\text{BWT}_{t}$ $\leq 0$ in most situations.

Learning \emph{plasticity} is also a crucial factor to consider. It refers to the model's capacity to effectively acquire new knowledge. Emphasizing too much on plasticity may lead to catastrophic forgetting, compromising the model's ability to retain previously learned tasks. On the other hand, excessively prioritizing stability may hinder the model's adaptability to new tasks. This is often referred to in the literature as the stability-plasticity dilemma~\cite{mermillod2013stability}. We quantify plasticity via the \textbf{intransigence measure (IM)}~\cite{chaudhry2018riemannian}, defined as
\begin{equation}
\text{IM}_{t} = \text{WER}_{t,t} - {\text{WER}_t^{\text{joint}}} \, \, \, t = 2, \dots, T, \label{eq:im}
\end{equation}
where the reference value ${\text{WER}_t^{\text{joint}}}$ denotes the word error rate on the $t$-th task of the model jointly trained on all tasks (\ie base $+$ new languages) at the same time.
{Note that} $\text{IM}_{t}$ $\in (-\infty, \infty)$ with larger values corresponding to the inability of the model to effectively learn new tasks. Also note that the choice of the reference value is arbitrary, however joint training is the most natural option as it provides an upper bound for the overall performance.

Lastly, we are interested in examining the \emph{influence} of previous tasks on learning the current task. While this concept is partly related to plasticity, it is worth noting that a model with large plasticity does not necessarily utilize knowledge from previous tasks to enhance performance on the current task. To capture this aspect, we propose a variation of \textbf{forward transfer (FWT)}~\cite{lopez2017gradient}, that we define as
\begin{equation}
\text{FWT}_{t} = {\text{WER}_t^{\text{fine-tuned}}} - \text{WER}_{t,t} \, \, \, t = 2, \dots, T, \label{eq:fwt}
\end{equation}
where ${\text{WER}_t^{\text{fine-tuned}}}$ denotes the word error rate on the $t$-th task of the model fine-tuned solely on that specific task.
{Note that} $\text{FWT}_{t}$ $\in (-\infty, \infty)$ with larger values corresponding to a stronger ability of the model to exploit knowledge from previous tasks.
Also note that our definition significantly deviates from the one of Lopez-Paz \etal~\cite{lopez2017gradient}, as they interpret forward, transfer as 
the zero-shot improvement in accuracy on future tasks compared to random guessing. However, in our setting, measuring performance on unknown tasks is not meaningful, as zero-shot transfer is difficult for multilingual ASR.

\section{Experiments}
\label{sec:experiments}
To show the utility of CL-MASR, we conduct a comparative study to determine the most effective combinations of CL methods and base models for multilingual ASR.
Additionally, we analyze the impact of language ordering.
All the experiments were run on 5 CentOS Linux machines with an Intel(R) Xeon(R) Silver 4216 Cascade Lake CPU with 32 cores @ 2.10 GHz, 64 GB RAM and an NVIDIA Tesla V100 SXM2 @ 32 GB with CUDA Toolkit 11.4.
With the specified hardware configuration, approximately 10 days are necessary to complete all the experiments.
For detailed information about the hyperparameters used in each experiment, refer to the official code repository.

\begin{figure*}[!t]
    \centering
    \begin{subfloat}{}
    \includegraphics[width=6.8in]{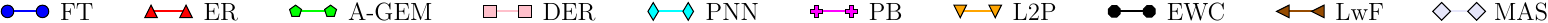}
    \end{subfloat}
    \begin{subfloat}{}
    \includegraphics[width=1.67in]{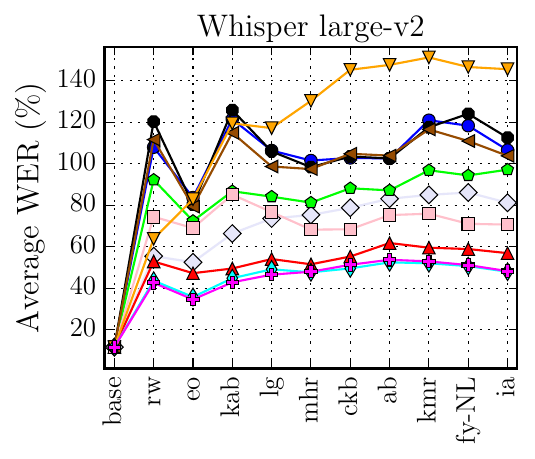}
    \end{subfloat}
    \vspace{-1.5mm}
    \begin{subfloat}{}
    \includegraphics[width=1.67in]{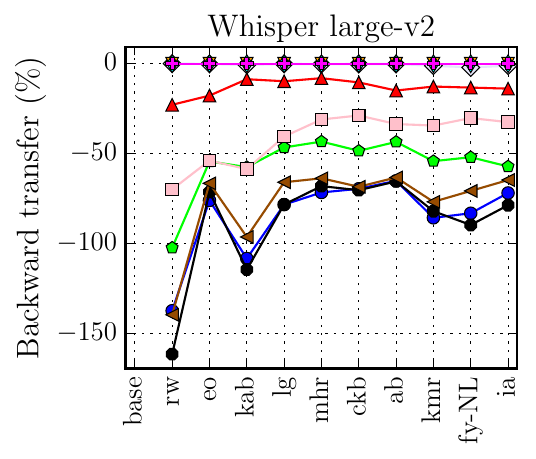}
    \end{subfloat}
    \begin{subfloat}{}
    \includegraphics[width=1.67in]{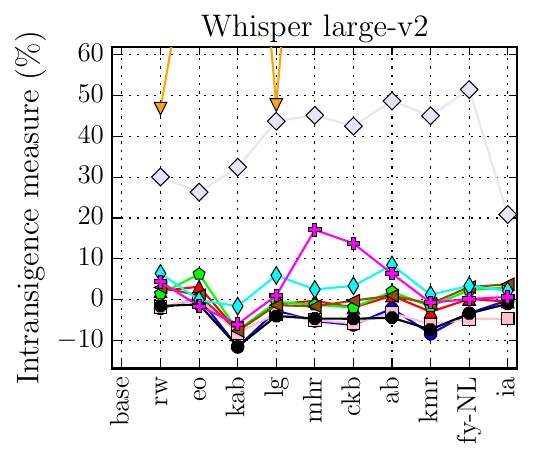}
    \end{subfloat}
    \begin{subfloat}{}
    \includegraphics[width=1.67in]{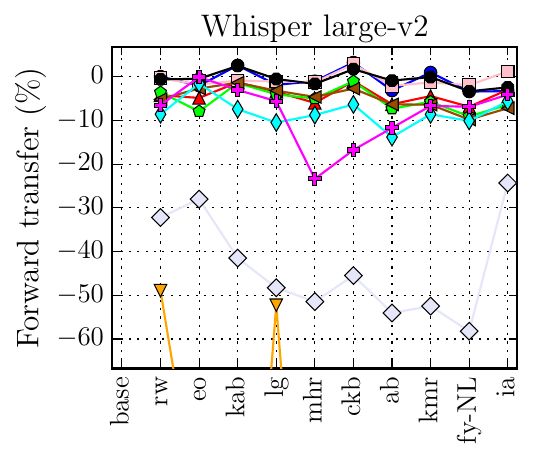}
    \end{subfloat}
    \centering
    \begin{subfloat}{}
    \includegraphics[width=1.67in]{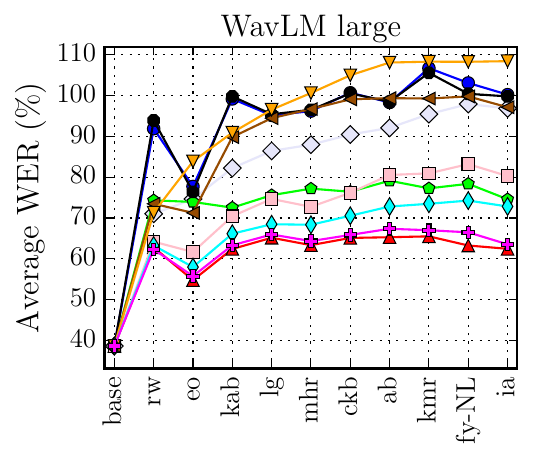}
    \end{subfloat}
    \begin{subfloat}{}
    \includegraphics[width=1.67in]{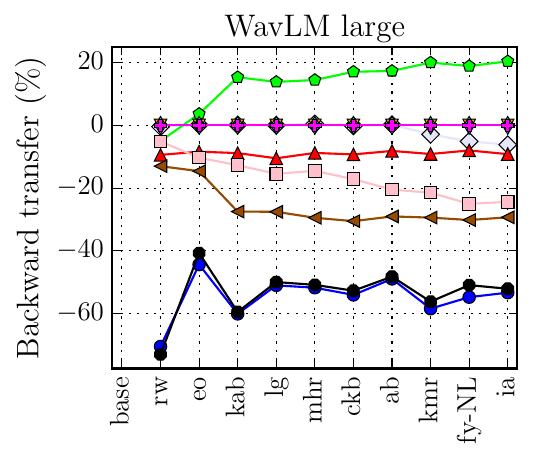}
    \end{subfloat}
    \begin{subfloat}{}
    \includegraphics[width=1.67in]{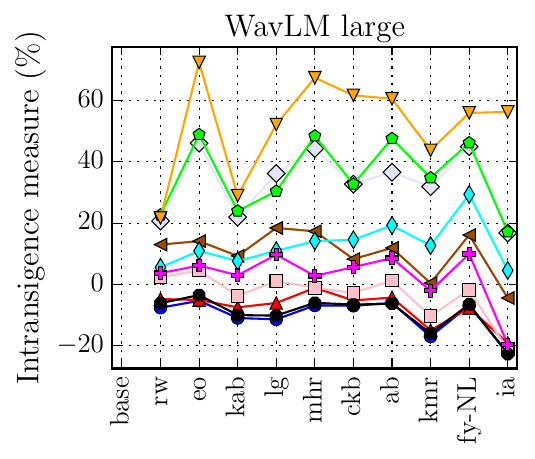}
    \end{subfloat}
    \begin{subfloat}{}
    \includegraphics[width=1.67in]{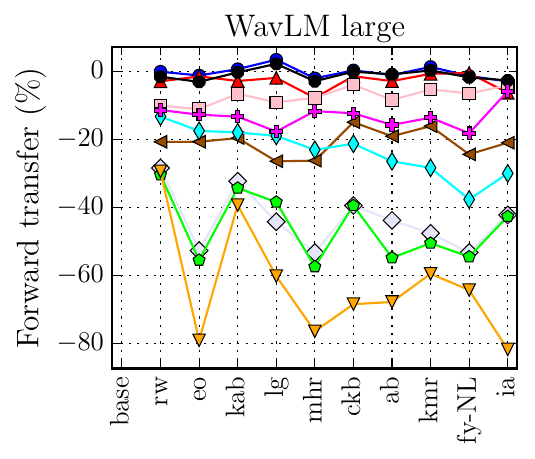}
    \end{subfloat}
\caption{Comparison of
fine-tuning (FT),
experience replay (ER), averaged gradient episodic memory (A-GEM), dark experience replay (DER), progressive neural networks (PNN), Piggyback (PB), learning to prompt (L2P), elastic weight consolidation (EWC), learning without forgetting (LwF), and memory aware synapses (MAS) applied to Whisper large-v2 and WavLM large on the base and new languages on Common Voice dataset. {Note that we crop the top right figures as L2P is off-scale. ER is among the best performing methods together with PNN and PB, with the additional advantage of being task-agnostic}.}
\label{fig:results}
\end{figure*}

\subsection{Comparison of Continual Learning Methods}
\cref{fig:results} shows the main results of our comparative study.
First of all, we observe that, when applying naive FT, the AWER is $\sim$100\% after learning each new language, which means that the model can hardly retain any knowledge about previous tasks {if no CL intervention is made}. This highlights the challenging nature of our benchmark, which makes it well-suited for the development of robust CL techniques.
Among the evaluated CL methods, two architecture-based approaches, namely PNN and PB, obtain the smallest AWER. Notably, they perform well despite the addition of \emph{small} adapter layers. This emphasizes the benefits of using large-scale pretrained models in CL.
However, L2P, which is also architecture-based, performs poorly. We hypothesize this is due to the complex nature of the sequence-to-sequence mapping task, which is significantly more challenging than the original task L2P was designed for, \ie image classification.
While ER performs comparably to PNN and PB, we would like to emphasize that it has the advantage of being task-agnostic at inference time, making it a more robust choice for multilingual ASR.
DER is also competitive to some degree, but the trade-off between forgetting and learning is too biased towards the latter, leading to an overall performance that is inferior to ER.
Finally, the regularization-based methods, namely EWC, LwF, and MAS, exhibit larger AWER, in most cases similar to naive FT.
Interestingly, MAS excels at mitigating forgetting but struggles with learning, possibly due to an overly strong regularization effect introduced by the parameter importance penalty.

Regarding the stability-plasticity trade-off, most methods are characterized by BWT $\le$ 0 and FWT $<$ 0. In particular, architecture-based strategies are immune to forgetting by design. However, this hinders their ability to learn new tasks, resulting in poor IM and FWT.
On the other hand, FT demonstrates superior adaptability to new tasks but falls short in mitigating forgetting.
These observations generally hold true for both base models.
The only exception is A-GEM, which yields to BWT $>0$ when applied to WavLM. 
Although not surprising (A-GEM explicitly aims to improve the model's performance on previous tasks), this effect is not observed for Whisper.
A possible explanation is that
Whisper's tokens are shared among languages, potentially introducing conflicts when optimizing A-GEM's loss function.
For additional experiments and more extensive results, refer to \cref{appendix:fleurs} and \cref{appendix:result}, respectively.

\subsection{Comparison of Base Models}
\label{subsec:base-models-comparison}
Based on \cref{fig:results}, we also draw a comparison between Whisper and WavLM.
We observe that Whisper tends to outperform WavLM in terms of AWER, especially on the base languages.
This is in line with expectations, as it was pretrained on a vast and diverse dataset encompassing 99 languages, unlike WavLM, which was pretrained on the 10 base languages only.
However, it is important to note that different stability-plasticity trade-offs are {achieved} by the two models.
Whisper exhibits better capacity in learning new languages, as indicated by smaller IM and larger FWT across all methods.
On the other hand, WavLM demonstrates superior performance in terms of BWT for both regularization-based and rehearsal-based methods, especially in mitigating forgetting for the base languages.
This difference can be attributed to several factors, including the pretraining strategies (supervised versus self-supervised) and, as previously noted, the token space (byte-level BPEs versus characters). In particular, we suspect that the type of tokens used in each model has an impact on the balance between learning and forgetting. To gain a deeper understanding and analyze these effects more comprehensively, further investigation is necessary.
Another interesting aspect is the steep increase in AWER and forgetting for both Whisper and WavLM after learning the first new language.
Our analysis suggests that this phenomenon is primarily associated with imbalance in terms of amount of data and/or number of training steps between the initial supervised joint pretraining and the subsequent incremental training. For more details, refer to \cref{appendix:imbalance}.

\subsection{Impact of Language Ordering}
\label{sec:ordering}

\begin{figure*}[t]
    \centering
    \begin{subfloat}{}
    \includegraphics[width=1.10in]{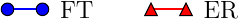}
    \end{subfloat}
    \\
    \begin{subfloat}{}
    \includegraphics[width=1.67in]{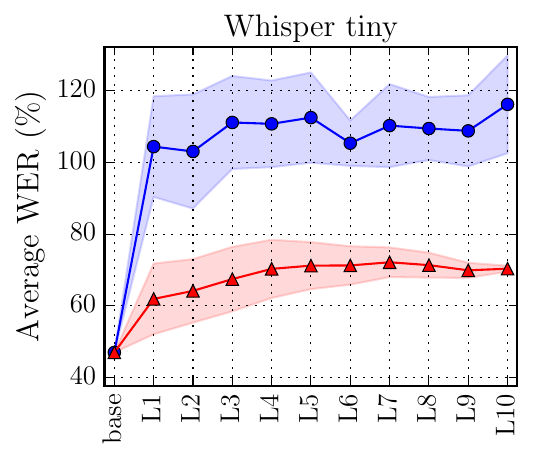}
    \end{subfloat}
    \begin{subfloat}{}
    \includegraphics[width=1.67in]{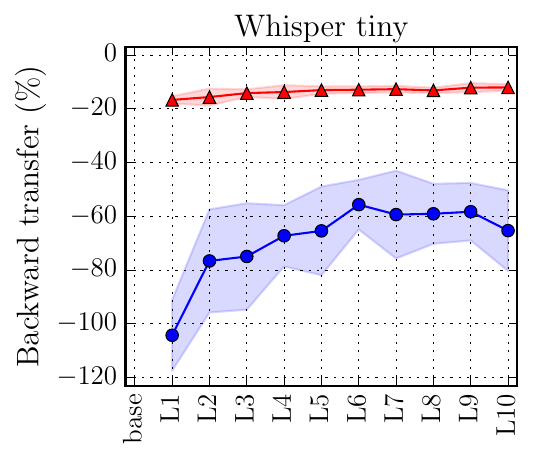}
    \end{subfloat}
    \begin{subfloat}{}
    \includegraphics[width=1.67in]{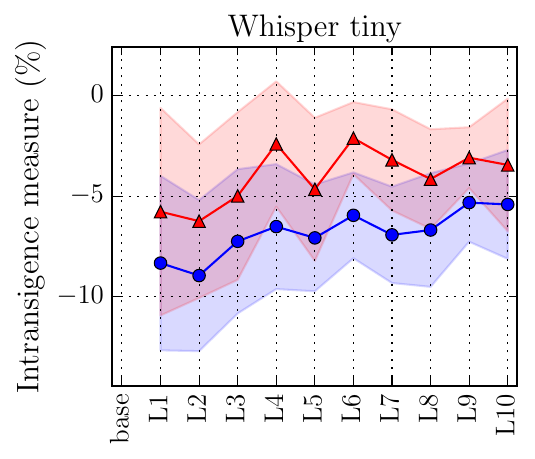}
    \end{subfloat}
    \begin{subfloat}{}
    \includegraphics[width=1.67in]{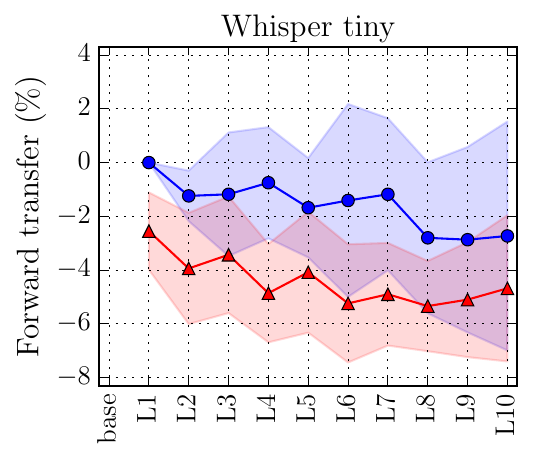}
    \end{subfloat}
\caption{Comparison of fine-tuning (FT) and experience replay (ER) applied to Whisper tiny. Curves represent the mean $\pm$ 1 standard deviation over 10 random language (L) orderings.
ER is effective in mitigating forgetting and reducing ordering sensitivity.}
\label{fig:ordering}
\end{figure*}

Performance variability due to task ordering is a primary concern in CL studies.
In \cref{fig:ordering}, we examine the impact of language ordering. Due to resource constraints, we limit our analysis to the tiny version of Whisper\footnote{\href{https://huggingface.co/openai/whisper-tiny}{https://huggingface.co/openai/whisper-tiny}}~\cite{radford2022robust} and we consider only FT and ER methods.
We observe that FT exhibits a significant fluctuation of 10-20\% in both AWER and BWT, depending on the language sequence.
On the other hand, ER is effective in mitigating forgetting and reducing ordering sensitivity: as more languages are added, the variance of AWER decreases, since the ordering becomes less important when a fraction of data from all previous tasks is accessible.
However, the ordering of languages still introduces substantial variance with respect to FWT and IM.
This highlights the importance of task ordering when transferring knowledge from previously learned languages to a new one.

\section{Conclusion}
\label{sec:conclusion}
We introduce CL-MASR, the first benchmark for continual learning in multilingual ASR.
CL-MASR offers a curated selection of medium/low-resource languages, a modular and flexible platform for executing and evaluating various CL methods on top of existing large-scale pretrained multilingual ASR models, and a standardized set of evaluation metrics.
Through a comparative study, we show that experience replay is one of the most effective strategies for combating catastrophic forgetting in multilingual ASR.
By releasing the code, we hope that CL-MASR will promote further research in the field and serve as a valuable real-world testbed for novel CL algorithms.

As a future work, we are planning to further explore other categories of CL
such as continual self-supervised representation learning~\cite{gallardo2021self,madaan2022representational,rao2019continual}. Additionally, deploying large-scale pretrained models poses challenges in practical applications due to the high computational burden. Using knowledge distillation~\cite{hsieh2023distilling,liang2020mixkd} to transfer knowledge from larger teacher models to smaller and more memory\nobreakdash-efficient student models as an alternative for base models could open up interesting directions for future work.

\section*{Acknowledgements}
We thank Reza Davari for valuable discussions.
We acknowledge the support of the Natural Sciences and Engineering Research Council of Canada (NSERC) and the Digital Research Alliance of Canada (alliancecan.ca).

\bibliographystyle{IEEEtran}

% \bibliography{bibliography}

\appendix
\subsection{General Information}
\label{appendix:info}

\subsubsection{Dataset Documentation}
The Common Voice 13~\cite{ardila2020common} is an openly accessible speech dataset that collects the voices of volunteers from all over the world. For extensive documentation refer to the official website (\url{https://commonvoice.mozilla.org/en/about}).
\newline
\subsubsection{Intended Uses}
CL-MASR is intended for researchers in continual learning and related fields in order for them to easily test their novel methods on a challenging real-word task.
\newline
\subsubsection{Hosting and Maintenance Plan}
CL-MASR platform is hosted and version-tracked via GitHub.
It is available at \url{https://github.com/speechbrain/benchmarks}.
The download link for the Common Voice 13~\cite{ardila2020common} dataset can be found on the official website (\url{https://commonvoice.mozilla.org/en/datasets}).

CL-MASR is a community-driven and open-source initiative. 
We plan to extend it by running additional experiments and including new continual learning methods and base models. We welcome external contributors.
\newline
\subsubsection{Licensing}
Our work is licensed under Apache 2.0 (\url{https://www.apache.org/licenses/LICENSE-2.0}).
The Common Voice 13~\cite{ardila2020common} dataset is licensed under CC0 public domain Creative Commons (\url{https://creativecommons.org/share-your-work/public-domain/cc0}).
\newline
\subsubsection{Author Statement}
We, the authors, will bear all responsibility in case of violation of rights.
\newline
\subsection{FLEURS Dataset}
\label{appendix:fleurs}

\begin{figure*}[!t]
    \centering
    \begin{subfloat}{}
    \includegraphics[width=6.8in]{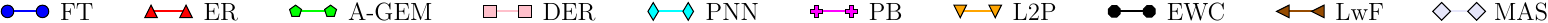}
    \end{subfloat}
    \begin{subfloat}{}
    \includegraphics[width=1.67in]{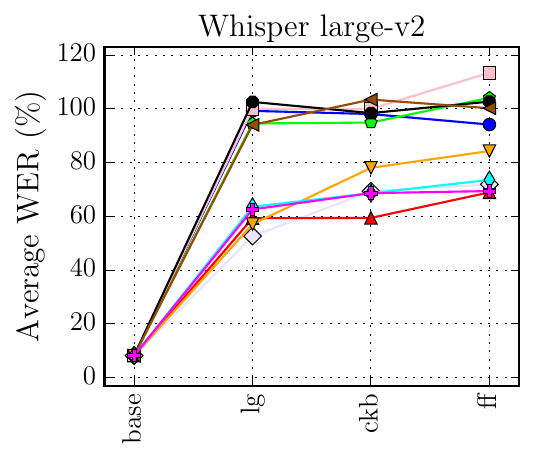}
    \end{subfloat}
    \vspace{-1.5mm}
    \begin{subfloat}{}
    \includegraphics[width=1.67in]{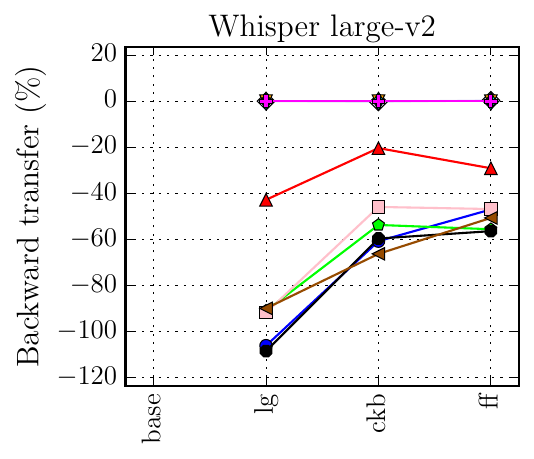}
    \end{subfloat}
    \begin{subfloat}{}
    \includegraphics[width=1.67in]{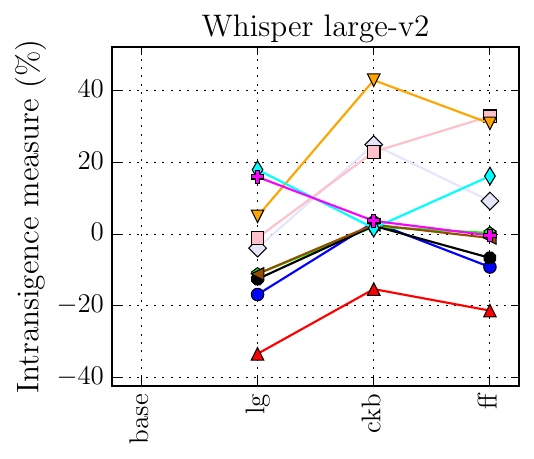}
    \end{subfloat}
    \begin{subfloat}{}
    \includegraphics[width=1.67in]{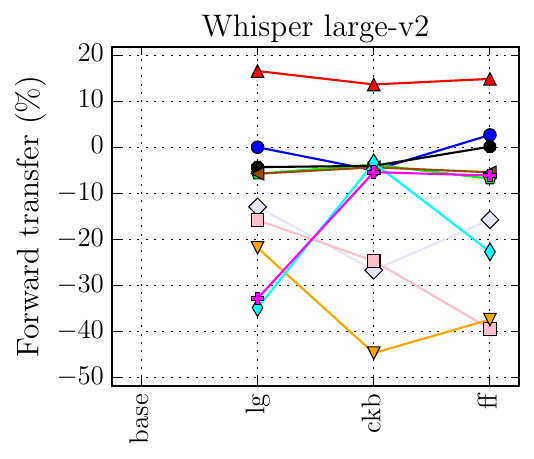}
    \end{subfloat}
    \centering
    \begin{subfloat}{}
    \includegraphics[width=1.67in]{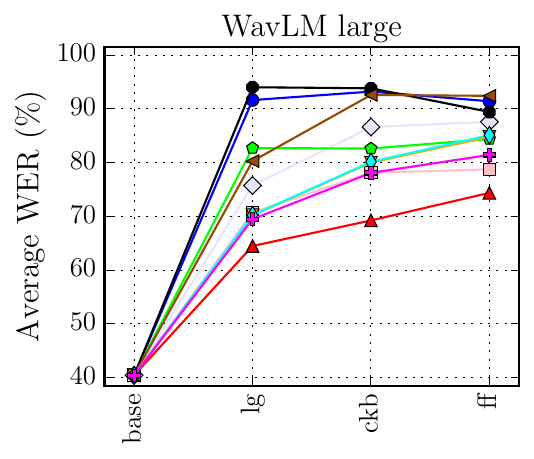}
    \end{subfloat}
    \begin{subfloat}{}
    \includegraphics[width=1.67in]{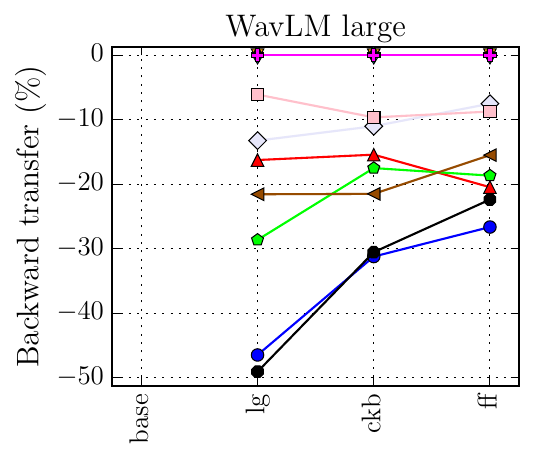}
    \end{subfloat}
    \begin{subfloat}{}
    \includegraphics[width=1.67in]{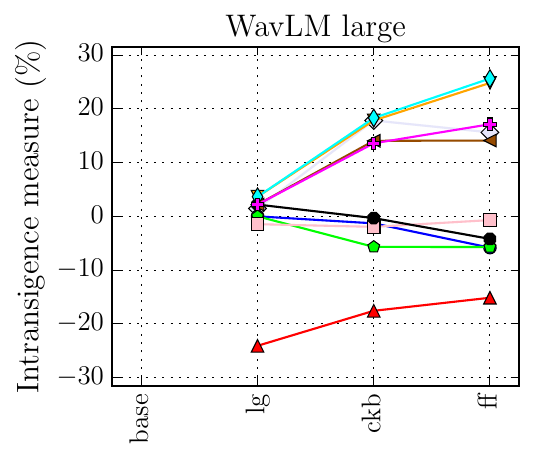}
    \end{subfloat}
    \begin{subfloat}{}
    \includegraphics[width=1.67in]{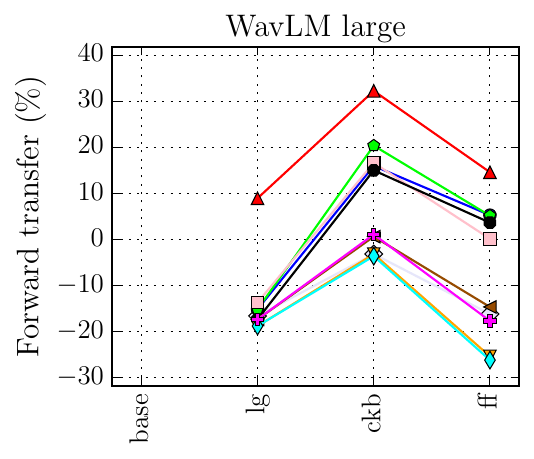}
    \end{subfloat}
\caption{{Comparison of fine-tuning (FT), experience replay (ER), averaged gradient episodic memory (A-GEM), dark experience replay (DER), progressive neural networks (PNN), Piggyback (PB), {learning to prompt (L2P)}, elastic weight consolidation (EWC), learning without forgetting (LwF), and memory aware synapses (MAS) applied to Whisper large-v2 and WavLM large on the base and new languages. ER achieves the smallest AWER across both Whisper and WavLM.}}
\label{fig:fleurs}
\end{figure*}

To showcase the flexibility of our benchmark platform and to further validate the results of our analysis, we extend our comparative study to the \textbf{F}ew-shot \textbf{L}earning \textbf{E}valuation of \textbf{U}niversal \textbf{R}epresentations of \textbf{S}peech (FLEURS)\footnote{\href{https://huggingface.co/datasets/google/xtreme_s}{https://huggingface.co/datasets/google/xtreme\_s}}\footnote{Licensed under CC-BY public domain Creative Commons (\url{https://creativecommons.org/licenses/by/4.0/}).}~\cite{fleurs} dataset. Derived from the machine translation FLoRes-101~\cite{goyal2022flores} benchmark, it consists of $n$-way parallel natural human speech and text in 102 languages, with approximately 12 hours of annotated data per language divided into training, validation, and test splits. Following the same procedure as Common Voice 13, we select from it two sets of languages ($L_\text{base} = L_\text{new} = 3$):
\begin{itemize}
    \item \textbf{base languages}: English (en), German (de), and Arabic (ar).
    \item \textbf{new languages}: Luganda (lg), Central Kurdish (ckb), and Fula (ff).
\end{itemize}
For each language, we randomly extract up to 10 hours of data for training, 1 for validation, and 1 for testing respectively. For the preprocessing and experimental evaluation, we use the same hyperparameters as Common Voice 13.% (see \cref{appendix: experiment}). 

\cref{fig:fleurs} shows the outcomes of our experiments on FLEURS.
The main findings align with those from Common Voice 13: rehearsal-based and architecture-based methods are generally effective for CL in ASR, while regularization-based approaches tend to perform poorly. Remarkably, ER achieves the smallest AWER across both Whisper and WavLM.
However, we observe some noteworthy exceptions. When applied to Whisper, DER struggles to both learn new tasks and mitigate forgetting. In contrast, MAS exhibits exceptional performance on Whisper, on par with ER. Lastly, when applied to WavLM, PNN shows scarce plasticity (WER $\sim$100\% for each new language), performing similarly to L2P.
This highlights how the performance of most CL methods is highly sensitive to both the dataset and the selection of hyperparameters.

\begin{figure}[!t]
    \centering
    \begin{subfloat}{}
    \includegraphics[width=3.45in]{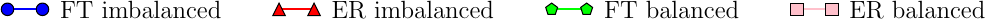}
    \end{subfloat}
    \\
    \begin{subfloat}{}
    \includegraphics[width=1.67in]{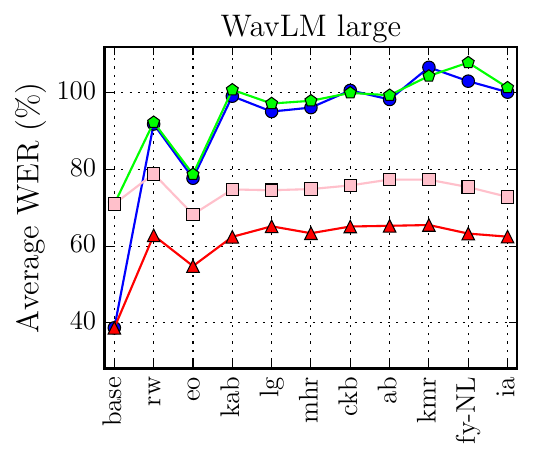}
    \end{subfloat}
    \begin{subfloat}{}
    \includegraphics[width=1.67in]{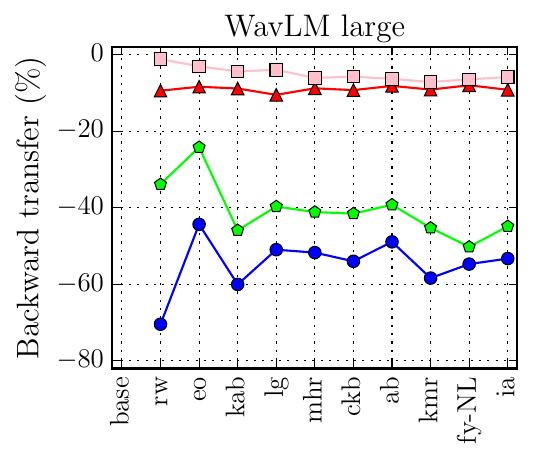}
    \end{subfloat}
\caption{{Comparison of the imbalanced and balanced variants of fine-tuning (FT) and experience replay (ER) applied to WavLM large on the base and new languages (10 hours per language). The steep increase in AWER and forgetting after learning the first new language is observed only for the imbalanced variants. ER effectively mitigates forgetting in both the balanced and imbalanced scenarios.}}
\label{fig:imbalance}
\end{figure}

\subsection{Effect of Imbalance between Base and New Languages}
\label{appendix:imbalance}
We hypothesize that the substantial drop in accuracy observed after learning the first new language (see \cref{fig:results}) is associated with imbalance in terms of amount of data and/or number of training steps between the initial supervised joint pretraining and the subsequent incremental training. To confirm this, we conduct an experiment involving imbalanced and balanced variants of fine-tuning (FT) and experience replay (ER) applied to WavLM large on the base and new languages (10 hours per language). Imbalanced means initial supervised joint pretraining on the base languages for 20 epochs, followed by incremental training for 2 epochs per language (as in the main results). Balanced means initial supervised joint pretraining on the base languages for 2 epochs, followed by incremental training for 2 epochs per language. \cref{fig:imbalance} shows that in the balanced scenario, the increase in forgetting is significantly smaller. This validates our hypothesis that imbalance is the primary cause of the observed phenomenon.
Even though the imbalanced case is characterized by this spike in WER, overall, AWER is smaller compared to the balanced case. This is expected, as the imbalanced case is trained with more data and/or epochs, and is likely the practical use case scenario.
Remarkably, ER effectively mitigates forgetting in both balanced and imbalanced scenarios.
\newpage
\subsection{Additional Results}\label{appendix:result}
Tables \ref{tab-awer}, \ref{tab-bwt}, \ref{tab-fwt}, and \ref{tab-im} present a more comprehensive results from our experiments on the Common Voice dataset.

\begin{table*}[!h]
\caption{The \textbf{average word error rate (AWER)} of fine-tuning (FT),
experience replay (ER), averaged gradient episodic memory (A-GEM), dark experience replay (DER), progressive neural networks (PNN), Piggyback (PB), learning to prompt (L2P), elastic weight consolidation (EWC), learning without forgetting (LwF), and memory aware synapses (MAS). For Whisper tiny, we report the mean $\pm$ 1 standard deviation over 10 random language (L) orderings. Smaller is better. The best value for each language is highlighted in bold.}
%\small
  \label{tab-awer}
  \centering
  \begin{tabular}{lcccccccccc}
    \toprule
    Language & FT & ER &  A-GEM & DER & PNN & PB & L2P & EWC & LwF & MAS \\
    \midrule
    \multicolumn{11}{c}{Whisper large-v2}\\
    \midrule
     base	& \textbf{11.63} &	\textbf{11.63} & 	\textbf{11.63} &	\textbf{11.63} & \textbf{11.63} &	\textbf{11.63} & \textbf{11.63} &	\textbf{11.63} &	\textbf{11.63} & \textbf{11.63} \\
    rw	& 107.92 &	52.88 &	 92.22 &	74.41 & 43.64 &	\textbf{42.60} & 63.82 &	120.28 &	111.64 & 55.43 \\
    eo	& 83.67 &	47.24 &	 72.40 &	68.98 & 35.78 &	\textbf{34.59} & 82.88 &	80.25 &	79.30 & 52.52 \\
    kab	& 121.42 &	49.50 &	 86.72 &	85.08 & 44.82 &	\textbf{42.84} & 119.41 &	125.79 &	114.88 & 66.35 \\
    lg	& 106.34 &	54.06 &	 84.10 &	76.68 & 49.11 &	\textbf{46.54} & 117.13 &	105.88 &	98.64 & 73.57 \\
    mhr	& 101.50 &	51.53 &	 81.22 &	68.22 & \textbf{47.66} &	47.95 & 130.27 &	98.30 &	97.40 & 75.30 \\
    ckb	& 102.77 &	55.27 &	 88.15 &	68.38 & \textbf{49.55} &	51.29 & 145.23 &	103.45 &	104.78 & 78.76 \\
    ab	& 102.52 &	61.77 &	 87.12 &	75.22 & \textbf{52.46} &	53.71 & 147.58 &	102.52 &	103.88 & 83.10 \\
    kmr	& 120.93 &	59.52 &	 96.83 &	75.91 & \textbf{52.02} &	52.92 & 151.18 &	117.54 &	116.50 & 84.96 \\
    fy-NL	& 118.29 &	58.87  &  94.33 &	71.03 & \textbf{50.65} & 51.14 &	146.56  & 124.01 &	110.90 & 86.11 \\
    ia	& 106.53 &	56.90 &	 97.12 &	70.65 & \textbf{47.95} &	48.25 & 145.51 &	112.51 &	103.78 & 81.17 \\
    \midrule
    {average}	& 98.50 &	50.83 &	 81.08 &	67.84 & 44.12 & 	\textbf{43.95} &	114.65 & 100.20 &	95.76 & 68.08 \\

     \midrule
     \multicolumn{11}{c}{WavLM large}\\
     
    \midrule
     base	& \textbf{38.63} &	\textbf{38.63} &	 \textbf{38.63} & \textbf{38.63} &	\textbf{38.63} &	\textbf{38.63} & \textbf{38.63} &	\textbf{38.63} &	\textbf{38.63} & \textbf{38.63}
\\ rw	& 91.88 &	62.79 &	74.32 &	64.20 & 63.25 &	\textbf{62.31} & 71.32 &	93.89 &	73.47 & 71.03
\\ eo & 77.72 &	\textbf{54.77} &	 73.93 &	61.66 & 57.98 &	55.76 & 83.88 &	76.39  &	71.26 & 74.64
\\ kab	& 99.16 &	\textbf{62.41} &	 72.53 &	70.43 & {66.15} &	63.32 & 90.91 &	99.78 &	89.82 & 82.22
\\ lg	& 95.14 &	\textbf{65.16} &	 75.59 &	74.67 & 68.47 &	65.96 & 96.53 &	95.41 &	94.45 & 86.42
\\ mhr	& 96.18 &	\textbf{63.35} &	 77.19 &	72.73 & 68.34 &	64.34 & 100.61 &	96.49 &	96.69 & 87.96
\\ ckb	& 100.67 &	\textbf{65.11} &	 76.43 &	76.17 & 70.55 &	65.85 & 104.95 &	100.51 &	99.07 & 90.44
\\ ab	& 98.30 &	\textbf{65.29} &	 79.13 &	80.54 & 72.81 &	67.36 & 108.08 &	98.49 &	99.36 & 92.08
\\ kmr	& 106.70 &	\textbf{65.50} &	 77.26 &	80.93  & 73.48 &	67.00 & 108.29 &	105.55 &	99.26 & 95.45
\\ fy-NL	& 103.08 &	\textbf{63.26} &	 78.37 &	83.20 & 74.27 &	66.49 & 108.26 &	100.43 &	99.78 & 97.91
\\ ia	& 100.21 &	\textbf{62.45} &	 74.59 & 80.27 & 72.82 &	63.55 & 108.42 &	99.78 &	97.13 & 96.85

\\ \midrule {average}	& 91.61 &	\textbf{60.79} &	 72.54 &	71.22 & 66.07 & 	61.87 &	92.72 & 91.40 &	87.17 & 83.06 \\

 \midrule
     \multicolumn{11}{c}{Whisper tiny}\\
    \midrule

base &	\textbf{47.01$_{\pm \text{0.00}}$} &	\textbf{47.01$_{\pm \text{0.00}}$} &	-- &	-- &	-- &	-- &	-- & -- & -- & --
\\ L1 &	104.46$_{\pm \text{14.06}}$ &	\textbf{61.94$_{\pm \text{9.87}}$} &	-- &	-- &	-- &	-- &	-- & -- & -- & --
\\ L2 &	103.10$_{\pm \text{15.92}}$ &	\textbf{64.19$_{\pm \text{8.87}}$} &	-- &	-- &	-- &	-- &	-- & -- & -- & --
\\ L3 &	111.20$_{\pm \text{12.98}}$ &	\textbf{67.48$_{\pm \text{9.00}}$} &	-- &	-- &	-- &	-- &	-- & -- & -- & --
\\ L4 &	110.81$_{\pm \text{12.09}}$ &	\textbf{70.33$_{\pm \text{8.13}}$} &	-- &	-- &	-- &	-- &	-- & -- & -- & --
\\ L5 &	112.56$_{\pm \text{12.60}}$ &	\textbf{71.24$_{\pm \text{6.57}}$} &	-- &	-- &	-- &	-- &	-- & -- & -- & --
\\ L6 &	105.41$_{\pm \text{6.34}}$ &	\textbf{71.29$_{\pm \text{5.36}}$} &	-- &	-- &	-- &	-- &	-- & -- & -- & --
\\ L7 &	110.32$_{\pm \text{11.59}}$ &	\textbf{72.18$_{\pm \text{4.14}}$} &	-- &	-- &	-- &	-- &	-- & -- & -- & --
\\ L8 &	109.51$_{\pm \text{8.80}}$ &	\textbf{71.38$_{\pm \text{3.48}}$} &	-- &	-- &	-- &	-- &	-- & -- & -- & --
\\ L9 &	108.85$_{\pm \text{9.93}}$ &	\textbf{69.93$_{\pm \text{2.11}}$} &	-- &	-- &	-- &	-- &	-- & -- & -- & --
\\ L10 &	116.25$_{\pm \text{13.66}}$ &	\textbf{70.38$_{\pm \text{0.82}}$} &	-- &	-- &	-- &	-- &	-- & -- & -- & --
\\ \midrule average &	103.59$_{\pm \text{3.48}}$ &	\textbf{67.03$_{\pm \text{1.88}}$} &	-- &	-- &	-- & -- &	-- &	--  & -- & -- \\

    \bottomrule
  \end{tabular}
\end{table*}

\newpage
\begin{table*}[!t]
\caption{The \textbf{backward transfer (BWT)} of fine-tuning (FT),
experience replay (ER), averaged gradient episodic memory (A-GEM), dark experience replay (DER), progressive neural networks (PNN), Piggyback (PB), learning to prompt (L2P), elastic weight consolidation (EWC), learning without forgetting (LwF), and memory aware synapses (MAS). For Whisper tiny, we report the mean $\pm$ 1 standard deviation over 10 random language (L) orderings. Larger is better. The best value for each language is highlighted in bold.}
%\small
  \label{tab-bwt}
  \centering
  \begin{tabular}{lcccccccccc}
    \toprule
    Language & FT & ER &  A-GEM & DER & PNN & PB & L2P & EWC & LwF & MAS \\
    \midrule
    \multicolumn{11}{c}{Whisper large-v2}\\
    \midrule

base &	-- & -- & -- & -- & -- & -- & -- & -- & -- & --
\\ rw &	-137.17 & 	-22.81 & 	-102.18 & -69.79 & 	\textbf{0.00} & 	\textbf{0.00} & \textbf{0.00} & 	-161.27 & 	-139.43 & 0.02
\\ eo &	-75.91 & 	-17.72 & 	-54.26 & -53.81 & 	\textbf{0.00} & 	\textbf{0.00} & \textbf{0.00}  & 	-71.24 & 	-66.43 & -0.10
\\ kab &	-108.17 & 	-8.62 & 	-57.42 & -58.53 & 	\textbf{0.00} & 	\textbf{0.00} & \textbf{0.00} & 	-114.28 & 	-96.19 & -0.68
\\ lg &	-78.24 & 	-9.75 & 	-46.52 & -40.46 & 	\textbf{0.00} & 	\textbf{0.00} & \textbf{0.00} & 	-78.19 & 	-65.83 & -0.12 
\\ mhr &	-71.50 & 	-8.04 & 	-43.24 & -30.96 & 	\textbf{0.00} & 	\textbf{0.00} & \textbf{0.00} & 	-67.98 & 	-63.63 & -0.27
\\ ckb &	-69.44 & 	-10.39 & 	-48.39 & -28.75 & 	\textbf{0.00} & 	\textbf{0.00} & \textbf{0.00} & 	-70.23 & 	-68.33 & -0.15
\\ ab &	-65.03 & 	-14.90 & 	-43.41 & -33.51 & 	\textbf{0.00} & 	\textbf{0.00} & \textbf{0.00} & 	-65.36 & 	-63.18 & -0.20 
\\ kmr &	-85.57 & 	-12.69 & 	-54.11 & -34.34 & 	\textbf{0.00} & 	\textbf{0.00} & \textbf{0.00} & 	-81.91 & 	-76.68 & -1.10 
\\ fy-NL &	-83.04 & 	-13.27 & 	-51.93 & -30.18 & 	\textbf{0.00} & 	\textbf{0.00} & \textbf{0.00} & 	-89.54 & 	-70.65 & -2.09
\\ ia &	-71.78 & 	-13.84 & 	-57.07 & -32.45 & 	\textbf{0.00} & 	\textbf{0.00} & \textbf{0.00} & 	-78.57 & 	-64.60 & -1.11
\\ \midrule average &	-84.58 & 	-13.20 & 	-55.85 & -41.28 & 	\textbf{0.00} & 	\textbf{0.00} & 	 \textbf{0.00} & -87.86 & 	-77.50 & -0.58  \\

     \midrule
     \multicolumn{11}{c}{WavLM large}\\
    \midrule
base &	-- &	-- &	-- &		-- &		-- &		-- &		-- & -- & -- & --
\\ rw &	-70.42 & -9.41 & -5.01 & -5.12 & \textbf{0.00} & \textbf{0.00} & \textbf{0.00} & -72.95 & -13.00 & -0.42
\\ eo &	-44.33 & -8.37 & \textbf{3.65} & -10.31 & 0.00 & 0.00 & {0.00} & -40.67 & -14.60 & 0.16
\\ kab &	-60.03 & -8.84 & \textbf{15.32} & -12.69 & 0.00 & 0.00 & {0.00} & -59.42 & -27.46 & 0.13
\\ lg &	-50.95 & -10.49 & \textbf{13.83} & -15.45 & 0.00 & 0.00 & {0.00} & -49.90 & -27.53 & 0.04
\\ mhr &	-51.71 & -8.78 & \textbf{14.44} & -14.49 & 0.00 & 0.00 & {0.00} & -50.80 & -29.43 & 0.48
\\ ckb &	-53.98 & -9.26 & \textbf{17.05} & -17.13 & 0.00 & 0.00 & {0.00} & -52.66  & -30.51 & -0.16
\\ ab &	-48.90 & -8.15 & \textbf{17.33} & -20.47 & 0.00 & 0.00 & {0.00} & -48.18 & -29.01 & 0.21
\\ kmr &	-58.37 & -9.12 & \textbf{20.00} & -21.44 & 0.00 & 0.00 & {0.00} & -56.13 & -29.37 & -2.86
\\ fy-NL &	-54.71 & -7.99 & \textbf{18.87} & -25.00 & 0.00 & 0.00 & {0.00} & -50.88 & -30.15 & -5.10
\\ ia &	-53.27 & -9.17 & \textbf{20.39} & -24.35 & 0.00 & 0.00 & {0.00} & -52.03 & -29.28 & -6.17
\\      \midrule
average &	-54.67 & -8.96 & \textbf{13.59} & -16.64 & 0.00 & 0.00 & {0.00} & -53.36 & -26.03  & -1.37 \\

 \midrule
     \multicolumn{11}{c}{Whisper tiny}\\
    \midrule

base &	-- &	-- &  -- &	-- &	-- &	-- &	-- & -- & -- & --
\\ L1 &	-104.27$_{\pm \text{13.20}}$	& \textbf{-16.68$_{\pm \text{1.36}}$}	&	-- &	-- &	-- &	-- &	-- & -- & -- & --
\\ L2 &	-76.60$_{\pm \text{19.20}}$		& \textbf{-15.61$_{\pm \text{3.14}}$} & -- &	-- &	-- &	-- &	-- & -- & -- & --
\\ L3 &	-74.94$_{\pm \text{19.81}}$		& \textbf{-14.14$_{\pm \text{1.41}}$} & -- &	-- &	-- &	-- &	-- & -- & -- & --
\\ L4 &	-67.24$_{\pm \text{11.41}}$		& \textbf{-13.75$_{\pm \text{2.56}}$} &	-- &	-- &	-- &	-- &	-- & -- & -- & --
\\ L5 &	-65.42$_{\pm \text{16.50}}$		& \textbf{-13.02$_{\pm \text{1.39}}$}	& -- &	-- &	-- &	-- &	-- & -- & -- & --
\\ L6 &	-55.66$_{\pm \text{9.24}}$		& \textbf{-12.88$_{\pm \text{1.22}}$}	& -- &	-- &	-- &	-- &	-- & -- & -- & --
\\ L7 &	-59.35$_{\pm \text{16.34}}$		& \textbf{-12.68$_{\pm \text{1.23}}$}	& -- &	-- &	-- &	-- &	-- & -- & -- & --
\\ L8 &	-59.06$_{\pm \text{11.18}}$		& \textbf{-13.15$_{\pm \text{1.03}}$}	& -- &	-- &	-- &	-- &	-- & -- & -- & --
\\ L9 &	-58.28$_{\pm \text{10.69}}$		& \textbf{-12.11$_{\pm \text{1.68}}$}	& -- &	-- &	-- &	-- &	-- & -- & -- & --
\\ L10 &	-65.32$_{\pm \text{15.01}}$	&	\textbf{-12.03$_{\pm \text{1.24}}$}	& -- &	-- &	-- &	-- &	-- & -- & -- & --
\\ \midrule
average &	-68.61$_{\pm \text{4.64}}$	&	\textbf{-13.61$_{\pm \text{0.55}}$} & -- &	-- &	-- &	-- &	-- & -- & -- & -- \\

    \bottomrule
  \end{tabular}
\end{table*}

\newpage
\begin{table*}[!t]
\caption{The \textbf{intransigence measure (IM)} of fine-tuning (FT), experience replay (ER), averaged gradient episodic memory (A-GEM), dark experience replay (DER), progressive neural networks (PNN), Piggyback (PB), learning to prompt (L2P), elastic weight consolidation (EWC), learning without forgetting (LwF), and memory aware synapses (MAS). For Whisper tiny, we report the mean $\pm$ 1 standard deviation over 10 random language (L) orderings. Smaller is better. The best value for each language is highlighted in bold.}
%\small
  \label{tab-im}
  \centering
  \begin{tabular}{lcccccccccc}
    \toprule
    Language & FT & ER &  A-GEM & DER & PNN & PB & L2P & EWC & LwF & MAS \\

    \midrule
    \multicolumn{11}{c}{Whisper large-v2}\\
    \midrule
base &	--	& --	& --	& --	& --	& --	& -- & -- & -- & --
\\ rw &	\textbf{-2.18} 	& 2.09 &	1.42 & -1.81 &	6.43 &	4.36 & 46.78 &	-1.55 &	3.00
& 30.03 \\ eo &	0.39 &	3.19 &	6.26 & 0.13  &	-0.08 &	\textbf{-1.58} & 100.86 &	-1.17 &	1.04 
& 26.34 \\ kab &	\textbf{-11.61} &	-7.73 &	-7.59 & -8.14 &	-1.61 &	-5.99 & 155.43 &	-11.51 &	-7.65
& 32.43 \\ lg &	{-2.70} &	-1.11 &	-0.53 & {-3.48} &	5.93 &	1.03 & 47.69 & \textbf{-4.03} &	-1.36
& 43.69 \\ mhr &	\textbf{-5.23} &	-0.28 &	-1.25 & -5.01 &	2.50 &	17.09 & 158.06 &	-4.62 &	-1.64
& 45.16 \\ ckb &	\textbf{-6.21} &	-1.95 &	-1.92 & -5.84 &	3.37 &	13.79 & 177.49 &	-4.69 &	-0.24
& 42.49 \\ ab &	{-2.18} &	0.96 &	1.98 & {-3.25} &	8.48 &	6.36 & 99.67 &	\textbf{-4.44} &	0.99
& 48.67 \\ kmr &	\textbf{-8.37} &	-2.98 &	-1.69 & -6.09 &	1.19 	& -0.74 & 132.72 &	-7.33 &	-1.01
& 45.04 \\ fy-NL &	-3.32 &	0.16 &	2.35 & \textbf{-4.79} &	3.38 	& 0.16 & 69.99 &	{-3.38} &	3.08
& 51.47 \\ ia &	-0.15 &	{-0.48} &	2.98 & \textbf{-4.66} &	2.17 &	0.66 & 116.30 &	-1.01 &	3.76 & 20.87
\\     \midrule
average &	{-4.16} &	-0.81 &	0.20 & \textbf{-4.29} &	3.18 &	3.51 & 110.50 &	-4.37 &	0.00 & 38.62 \\

     \midrule
     \multicolumn{11}{c}{WavLM large}\\
    \midrule
base &	--	& --	& --	& --	& --	& --	& -- & -- & -- & --
\\ rw &	\textbf{-7.64} & -4.80 &	22.66 & 2.31 &	5.53 &	3.64 & 21.66 &	-6.15  &	12.97 
& 20.66 \\ eo &	\textbf{-5.38} &	-5.12 &	48.92 & 4.53 &	10.91 &	6.14 & 72.47 &	-3.51 &	14.11 
& 46.09 \\ kab &	\textbf{-11.00} &	-7.52 &	23.95 & -3.76 &	7.59 &	2.96 & 28.95 &	-10.04 &	9.26 
& 21.95 \\ lg &	\textbf{-11.44} &	-6.09 &	30.41 & 1.09 &	10.97 &	9.68 & 52.19 &	-10.22 &	18.42 
& 36.19 \\ mhr &	\textbf{-6.89} &	-1.17 &	48.47 & -1.15 &	14.12 &	2.75 & 67.47 &	-6.04 &	17.31 
& 44.35 \\ ckb &	\textbf{-7.00} &	-5.27 &	32.73 & -2.78 &	14.57 &	5.62 & 61.73 &	-6.62 &	8.25 
& 32.73 \\ ab &	{-6.08} &	-4.31 &	47.61 & 1.23 &	19.25 &	8.57 & 60.61 &	\textbf{-6.31} &	11.88 
& 36.61 \\ kmr &	\textbf{-17.03} &	-15.00 &	34.78 & -10.40 &	12.64 &	-2.11 & 43.78 &	-16.00 &	0.38 
& 31.88 \\ fy-NL &	-6.91 &	\textbf{-7.78} &	46.22 & -1.81 &	29.37 &	9.91 & 56.00 &	-6.52 &	16.11 
& 44.95 \\ ia &	{-22.50} &	-19.16 &	17.18 & -21.33 &	4.54 &	-19.62 & 56.29 &	\textbf{-22.77} &	-4.47 
& 16.75 \\ \midrule average &	\textbf{-10.19} &	-7.62 &	35.29 & -3.21 &	12.95 &	2.75 & 52.11 &	-9.42 &	10.42 & 33.22 \\

 \midrule
     \multicolumn{11}{c}{Whisper tiny}\\
    \midrule

    base &	-- &	-- &	-- &	-- &	-- &	-- &	-- &	-- &	-- &	--
\\ L1 &	\textbf{-8.32$_{\pm \text{4.33}}$} &	-5.77$_{\pm \text{5.15}}$ &	-- &	-- &	-- &	-- &	-- &	-- &	-- &	--
\\ L2 &	\textbf{-8.94$_{\pm \text{3.75}}$}	& -6.24$_{\pm \text{3.81}}$	& -- &	-- &	-- &	-- &	-- &	-- &	-- &	--
\\ L3 &	\textbf{-7.24$_{\pm \text{3.58}}$}	& -4.99$_{\pm \text{4.16}}$	& -- &	-- &	-- &	-- &	-- &	-- &	-- &	--
\\ L4 &	\textbf{-6.51$_{\pm \text{3.10}}$}	& -2.41$_{\pm \text{3.11}}$	& -- &	-- &	-- &	-- &	-- &	-- &	-- &	--
\\ L5 &	\textbf{-7.07$_{\pm \text{2.65}}$}	& -4.66$_{\pm \text{3.55}}$	& -- &	-- &	-- &	-- &	-- &	-- &	-- &	--
\\ L6 &	\textbf{-5.95$_{\pm \text{2.13}}$}	& -2.12$_{\pm \text{1.79}}$	& -- &	-- &	-- &	-- &	-- &	-- &	-- &	--
\\ L7 &	\textbf{-6.92$_{\pm \text{2.39}}$}	& -3.20$_{\pm \text{2.51}}$	& -- &	-- &	-- &	-- &	-- &	-- &	-- &	--
\\ L8 &	\textbf{-6.68$_{\pm \text{2.82}}$}	& -4.15$_{\pm \text{2.47}}$	& -- &	-- &	-- &	-- &	-- &	-- &	-- &	--
\\ L9 &	\textbf{-5.32$_{\pm \text{1.94}}$}	& -3.09$_{\pm \text{1.51}}$	& -- &	-- &	-- &	-- &	-- &	-- &	-- &	--
\\ L10 &	\textbf{-5.41$_{\pm \text{2.70}}$}	& -3.45$_{\pm \text{3.29}}$	& -- &	-- &	-- &	-- &	-- &	-- &	-- &	--
\\ \midrule average &	\textbf{-6.84$_{\pm \text{0.96}}$}	& -4.01$_{\pm \text{1.05}}$	& -- &	-- &	-- &	-- &	-- &	-- &	-- &	-- \\

    \bottomrule
  \end{tabular}
\end{table*}

\newpage
\begin{table*}[!t]
\caption{The \textbf{forward transfer (FWT)} of fine-tuning (FT), experience replay (ER), averaged gradient episodic memory (A-GEM), dark experience replay (DER), progressive neural networks (PNN), Piggyback (PB), learning to prompt (L2P), elastic weight consolidation (EWC), learning without forgetting (LwF), and memory aware synapses (MAS). For Whisper tiny, we report the mean $\pm$ 1 standard deviation over 10 random language (L) orderings. Larger is better. The best value for each language is highlighted in bold.}
%\small
  \label{tab-fwt}
  \centering
  \begin{tabular}{lcccccccccc}
    \toprule
    Language & FT & ER &  A-GEM & DER & PNN & PB & L2P & EWC & LwF & MAS  \\
    \midrule
    \multicolumn{11}{c}{Whisper large-v2}\\
    \midrule
base &	--	& --	& --	& --	& --	& --	& -- & --	& --	& --
\\ rw &	\textbf{0.00} &	-4.27 &	-3.60 & -0.37 &	-8.61 &	-6.54 & -48.96 &	-0.63 &	-5.18
& -32.21 \\ eo &	-2.08 &	-4.88 &	-7.95 & -1.82 &	-1.61 &	\textbf{-0.11} & -102.55 &	{-0.52} &	-2.73
& -28.03 \\ kab &	\textbf{2.55} &	-1.33 &	-1.47 & -0.92 &	-7.45 &	-3.07 & -164.49 &	2.45 &	-1.41
& -41.49 \\ lg &	{-1.89} &	-3.48 &	-4.06 & {-1.11} &	-10.52 &	-5.62 & -52.28 &	\textbf{-0.56} &	-3.23
& -48.28 \\ mhr &	\textbf{-1.07} &	-6.02 &	-5.05 & -1.29 &	-8.80 &	-23.39 & -164.36 &	-1.68 &	-4.66
& -51.46 \\ ckb &	\textbf{3.21} &	-1.05 &	-1.08 & 2.84 &	-6.37 &	-16.79 & -180.49  &	1.69 &	-2.76
& -45.49 \\ ab &	-3.19 &	-6.33 &	-7.35 & {-2.12} &	-13.85 &	-11.73 & -105.04 &	\textbf{-0.93} &	-6.36
& -54.04 \\ kmr &	\textbf{0.93} &	-4.46 &	-5.75 & -1.35 &	-8.63 &	-6.70 & -140.16 &	-0.11 &	-6.43
& -52.48 \\ fy-NL &	-3.43 &	-6.91 &	-9.10 & \textbf{-1.96} &	-10.13 &	-6.91 & -76.74 &	-3.37 &	-9.83
& -58.22 \\ ia &	-3.33 &	-3.00 &	-6.46 & \textbf{1.18} &	-5.65 &	-4.14 & -119.78 &	-2.47 &	-7.24
& -24.35 \\ \midrule average &	-0.83 &	-4.17 &	-5.19 & {-0.69} &	-8.16 &	-8.50 & -115.48 &	\textbf{-0.61} &	-4.98 & -43.61 \\

     \midrule
     \multicolumn{11}{c}{WavLM large}\\
    \midrule
base &	--	& --	& --	& --	& --	& --	& --  & --	& --	& --
\\ rw &	\textbf{0.00} &	-2.84 &	-30.30 & -9.95 &	-13.17 &	-11.28 & -29.30 &	-1.49  &	-20.61 & -28.30
\\ eo &	\textbf{-1.11} & -1.37 &	-55.41 & -11.02 &	-17.40 &	-12.63 & -78.96 &	-2.98 &	-20.60 & -52.58
\\ kab &	\textbf{0.75} &	-2.73 &	-34.20 & -6.49 &	-17.84 &	-13.21 & -39.20 &	-0.21 &	-19.51  & -32.20
\\ lg &	\textbf{3.53} &	-1.82 &	-38.32 & -9.00 &	-18.88 &	-17.59 & -60.10 &	2.31 &	-26.33  & -44.10
\\ mhr &	\textbf{-1.95} &	-7.67 &	-57.31 & -7.69 &	-22.96 &	-11.59 & -76.31 &	-2.80 &	-26.15  & -53.19
\\ ckb &	\textbf{0.40} &	-1.33 &	-39.33 & -3.82 &	-21.17 &	-12.22 & -68.33 &	0.02 &	-14.85 & -39.33
\\ ab &	{-1.00} &	-2.77 &	-54.69 & -8.31 &	-26.33 &	-15.65 & -67.69 &	\textbf{-0.77} &	-18.96 & -43.69
\\ kmr &	\textbf{1.41} &	-0.62 &	-50.40 & -5.22 &	-28.26 &	-13.51 & -59.40 &	0.38 &	-16.00 & -47.50
\\ fy-NL &	-1.27 &	\textbf{-0.40} &	-54.40 & -6.37 &	-37.55 &	-18.09 & -64.18 &	-1.66  &	-24.29 & -53.13
\\ ia &	-2.85 &	-6.19 &	-42.53 & -4.02 &	-29.89  & -5.73 & -81.64 &	\textbf{-2.58} &	-20.88 & -42.10
\\ \midrule average &	\textbf{-0.21} &	-2.77 &	-45.69 & -7.19 &	-23.34 &	-13.15 & -62.51 &	-0.98 &	-20.82 & -43.61 \\

 \midrule
     \multicolumn{11}{c}{Whisper tiny}\\
    \midrule

base &	-- &	-- &	-- &	-- &	-- &	-- &	-- & --	& --	& --
\\ L1 &	\textbf{0.00$_{\pm \text{0.00}}$} &	-2.55$_{\pm \text{1.44}}$ &		-- &	-- &	-- &	-- &	-- & -- & -- & --
\\ L2 &	\textbf{-1.24$_{\pm \text{0.94}}$} &		-3.94$_{\pm \text{2.07}}$ &		-- &	-- &	-- &	-- &	-- & --	& --	& --
\\ L3 &	\textbf{-1.18$_{\pm \text{2.30}}$} &		-3.43$_{\pm \text{2.17}}$ &		-- &	-- &	-- &	-- &	-- & --	& --	& --
\\ L4 &	\textbf{-0.75$_{\pm \text{2.07}}$}  &		-4.86$_{\pm \text{1.83}}$ &		-- &	-- &	-- &	-- &	-- & --	& --	& --
\\ L5 &	\textbf{-1.68$_{\pm \text{1.85}}$} &		-4.08$_{\pm \text{2.25}}$ &		-- &	-- &	-- &	-- &	-- & --	& --	& --
\\ L6 &	\textbf{-1.41$_{\pm \text{3.59}}$} &		-5.24$_{\pm \text{2.20}}$ &		-- &	-- &	-- &	-- &	-- & --	& --	& --
\\ L7 &	\textbf{-1.18$_{\pm \text{2.83}}$} &		-4.90$_{\pm \text{1.91}}$ &		-- &	-- &	-- &	-- &	-- & --	& --	& --
\\ L8 &	\textbf{-2.80$_{\pm \text{2.81}}$} &		-5.34$_{\pm \text{1.68}}$ &		-- &	-- &	-- &	-- &	-- & --	& --	& --
\\ L9 &	\textbf{-2.87$_{\pm \text{3.45}}$} &		-5.10$_{\pm \text{2.14}}$ &		-- &	-- &	-- &	-- &	-- & --	& --	& --
\\ L10 &	\textbf{-2.73$_{\pm \text{4.26}}$} &		-4.68$_{\pm \text{2.72}}$ &		-- &	-- &	-- &	-- &	-- & --	& --	& --
\\ \midrule average &	\textbf{-1.58$_{\pm \text{0.85}}$} &		-4.41$_{\pm \text{0.65}}$ &		-- &	-- &	-- &	-- &	--  & --	& --	& -- \\

    \bottomrule
  \end{tabular}
\end{table*}

\end{document}